\documentclass{article}

     \PassOptionsToPackage{numbers, compress}{natbib}
    \usepackage[preprint]{neurips_2024}

\usepackage[utf8]{inputenc} % allow utf-8 input
\usepackage[T1]{fontenc}    % use 8-bit T1 fonts
\usepackage{hyperref}       % hyperlinks
\usepackage{url}            % simple URL typesetting
\usepackage{booktabs}       % professional-quality tables
\usepackage{amsfonts}       % blackboard math symbols
\usepackage{nicefrac}       % compact symbols for 1/2, etc.
\usepackage{microtype}      % microtypography
\usepackage{xcolor}         % colors

\usepackage{subfigure}
\usepackage{graphicx}
\usepackage{hyperref}       %
\usepackage{url}            %
\usepackage{booktabs}       %
\usepackage{amsfonts}       %
\usepackage{amsmath}
\usepackage{amssymb}
\usepackage{amsthm}         %
\usepackage{algorithm}
\usepackage{algorithmic}
\usepackage{nicefrac}       %
\usepackage{microtype}  
\usepackage{makecell}    %
\usepackage {multirow}
\usepackage[figuresright]{rotating}
\definecolor{mine}{RGB}{205, 232, 248}%

\usepackage{hyperref}
\usepackage{bbm}

\usepackage{diagbox}
\usepackage{enumitem}
\usepackage[most]{tcolorbox}
\usepackage{wrapfig}

\usepackage{cleveref}
\crefname{equation}{Equation}{Equations}
\crefname{section}{Section}{Sections}
\crefname{figure}{Figure}{Figures}
\crefname{table}{Table}{Tables}
\crefname{appendix}{Appendix}{Appendix}
\crefname{theorem}{Theorem}{Theorems}

\theoremstyle{plain}

\theoremstyle{definition}

\theoremstyle{remark}

\newcount\Comments
\providecommand{\dataset}[0]{\mathcal{D}}
\providecommand{\promptdataset}[0]{\mathcal{D}_\mathcal{X}}

\def\bx{{\boldsymbol{x}} }
\def\by{{\boldsymbol{y}} }

\newcommand{\E}{\mathbb{E}}

\newcommand{\KL}{\mathrm{\mathbb D_{KL}}}

\title{Online Self-Preferring Language Models}

\author{%
  Yuanzhao Zhai\textsuperscript{\rm 1, \rm 2}, Zhuo Zhang\textsuperscript{\rm 3, \rm 4}, Kele Xu\textsuperscript{\rm 1, \rm 2}, Hanyang Peng\textsuperscript{\rm 4},  \\ \textbf{Yue Yu\textsuperscript{\rm 1, \rm 2},  Dawei Feng\textsuperscript{\rm 1, \rm 2}\thanks{davyfeng.c@gmail.com}, Cheng Yang\textsuperscript{\rm 1, \rm 2}, Bo Ding\textsuperscript{\rm 1, \rm 2}, Huaimin Wang\textsuperscript{\rm 1, \rm 2}} \\
    \textsuperscript{\rm 1}National University of Defense Technology, Changsha, China\\
    \textsuperscript{\rm 2}State Key Laboratory of Complex \& Critical Software Environment\\
    \textsuperscript{\rm 3}Harbin Institute of Technology (Shenzhen), Shenzhen, China\\
    \textsuperscript{\rm 4}Peng Cheng Laboratory, Shenzhen, China\\
}

\begin{document}

\maketitle

\begin{abstract}

Aligning with human preference datasets has been critical to the success of large language models (LLMs).
Reinforcement learning from human feedback (RLHF) employs a costly reward model to provide feedback for on-policy sampling responses.
Recently, offline methods that directly fit responses with binary preferences in the dataset have emerged as alternatives.
However, existing methods do not explicitly model preference strength information, which is crucial for distinguishing different response pairs.
To overcome this limitation, we propose Online Self-Preferring (OSP) language models to learn from self-generated response pairs and self-judged preference strengths.
For each prompt and corresponding self-generated responses, we introduce a ranked pairing method to construct multiple response pairs with preference strength information.
We then propose the soft-preference cross-entropy loss to leverage such information.
Empirically, we demonstrate that leveraging preference strength is crucial for avoiding overfitting and enhancing alignment performance. 
OSP achieves state-of-the-art alignment performance across various metrics in two widely used human preference datasets.
OSP is parameter-efficient and more robust than the dominant online method, RLHF when limited offline data are available and generalizing to out-of-domain tasks.
Moreover, OSP language models established by LLMs with proficiency in self-preferring can efficiently self-improve without external supervision.

\end{abstract}

\section{Introduction}
Fine-tuning large language models (LLMs) to align human preferences has gained significant attention in various natural language processing (NLP) tasks~\cite{2020GPT3,ji2023ai}.
Existing alignment methods can be broadly classified into online and offline, based on whether on-the-fly responses and feedback are collected~\cite{levine2020offline}.
The representative online method, RLHF~\cite{InstructGPT} first learns a computationally intensive reward model and then optimizes LLMs to generate responses with high rewards provided by reward models.
Offline methods that do not utilize newly collected responses and preferences, emerged as an efficient alternative to RLHF.
Notably, direct preference optimization (DPO)~\cite{DPO} and its variants~\cite{IPO, zhao2023sliHF} sidestep the reward modeling and directly fit binary preference labels.

% However, both substituted rewards and binary preferences fail to capture the preference strength information within response pairs.

% fail to express complex preference relations, such as non-transitive or cyclic preference~\cite{bertrand2023limitations,munos2023nash}.

% While successful, these approaches fail to capture the preference strength information within the response pairs.

Human preference datasets commonly consist of only binary preferences for response pairs.
Nevertheless, preference strength, the degree to which one response is preferred over another, is crucial information within these pairs.
However, such preference strength information cannot be fully utilized by substituted rewards~\cite{NLHF} or binary preferences.
In this work, we aim to leverage the preference strength of response pairs for alignment.
We employ the on-policy response sampling method to collect response pairs, improving the data coverage and quality over the offline dataset~\cite{lambert2022challenges, online-DPO,tang2024understanding}.
We then differentiate these pairs using soft preference strengths instead of treating them equivalently as binary preferences.
Moreover, inspired by recent success with self-feedback~\cite{yuan2024self,Self-judge}, we propose preferring response pairs using the same LLM that generated them, eliminating the need for a separate model and dramatically reducing computational overhead compared to RLHF.

% , can be provided via LLM-as-a-judge~\cite{zheng2023judging}.
% However, both substituted rewards and binary preferences fail to capture such complex preference strength information~\cite{NLHF}.

% , which is demonstrated to improve the data coverage and quality over the offline dataset~\cite{lambert2022challenges, online-DPO,tang2024understanding}.

% using on-policy sampling, aiming to improve the data coverage and quality over the offline dataset.
% With soft preference strength, we can distinguish multiple response pairs for one prompt, rather than treating all pairs with binary preference as equivalent.

% We refer to preference strength as the degree to how much one response is preferred over another.
% Such information can be provided by LLM-as-a-judge, with which we can distinguish different response pairs.
% , rather than treating all pairs with binary preferences as equivalent.

% and learn from soft preference strengths within the pairs.

% on-the-fly response pairs as well as corresponding soft preference strengths.
% Besides, inspired by recent progress on self-feedback~\cite{yuan2024self,Self-judge}, we propose preferring response pairs using the same LLM as used for generation, eliminating the need for a separate preference model and dramatically reducing computational overhead.

\begin{figure*}[t]
    % \vskip 0.2in
    \begin{center}
    \centerline{\includegraphics[width=0.93\columnwidth]{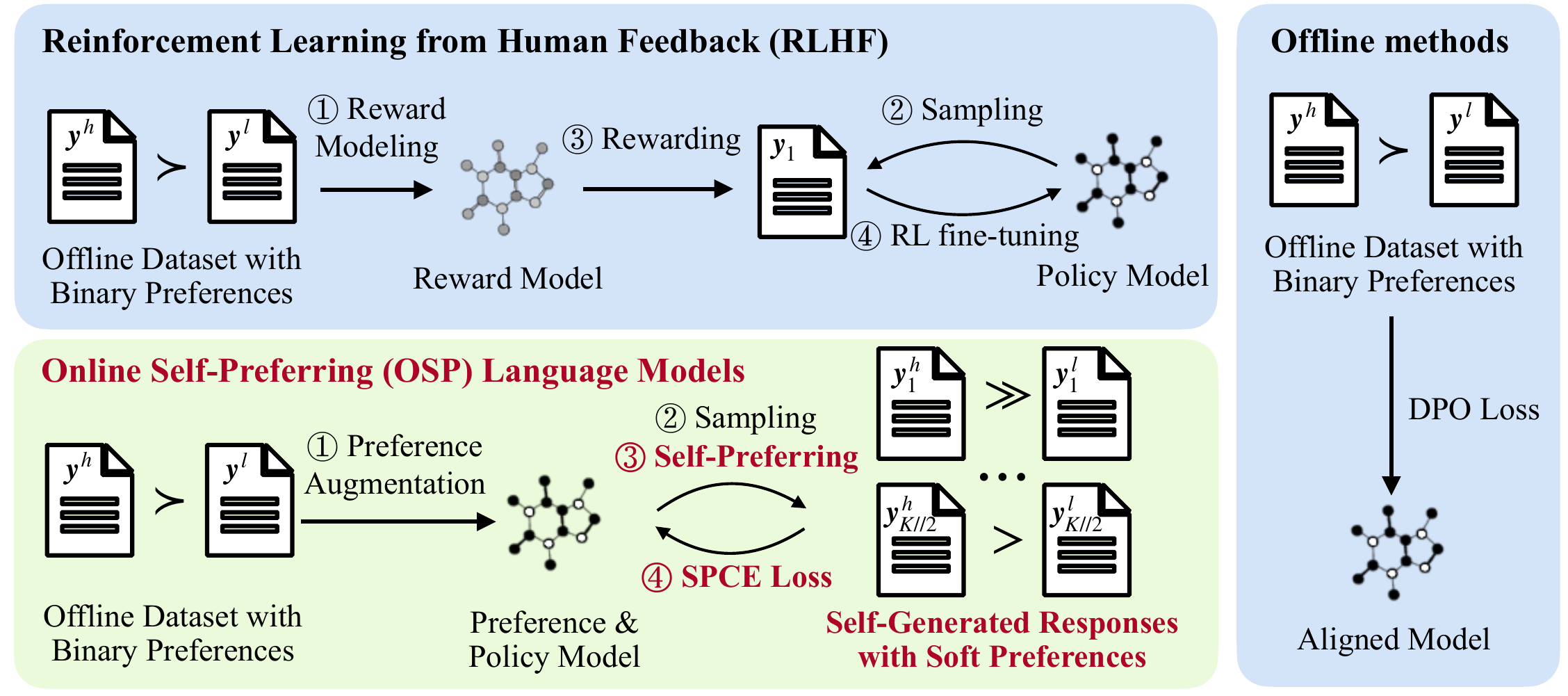}}
    \vspace{-0.5em}
    \caption{Comparison of LLMs alignment methods.
    Compared to RLHF, OSP leverages the LLM itself to provide the preference strength of response pairs, instead of leveraging a separate reward model to reward single responses. In contrast to offline methods, OSP can effectively learn from on-the-fly self-generated samples and their associated preference strengths.
    }
    \label{fig:comparison}
    \end{center}
    \vspace{-1em}
\end{figure*}

We thus introduce online self-preferring language models, which learn from multiple on-the-fly self-generated response pairs with various self-judged preference strengths, as illustrated in \cref{fig:comparison}.
The offline human preference dataset can be used for preference augmentation.
For each prompt, the same LLM first samples multiple response pairs and then judges the preference strength for each one over the others.
Since existing direct alignment loss is designed for the offline setting and does not account for preference strength, we further propose the soft-preference cross-entropy (SPCE) loss, which can mitigate overfitting and improve alignment performance.
Results on two widely used human preference datasets show that OSP outperforms existing online and offline alignment methods and remains effective with only $2\%$ of the datasets.
OSP language models can effectively generalize to OOD tasks, mitigating the reward hacking issue faced by RLHF.
Additionally, the OSP language model established by Mistral-7B-Instruct-v0.2, which is proficient in self-preferring, achieves win rates of $95\%$ against the original model in the summarization task using only $200$ prompts for self-improvement.
In summary, our main contributions are as follows: 
\begin{itemize}
    \item We introduce the online self-preferring language model for LLM alignment, which learns from self-generated response pairs and self-judged soft preference strength.
    \item We propose a soft-preference cross-entropy loss to effectively leverage multiple response pairs with different preference strengths.
    \item Empirical results demonstrate that our method outperforms state-of-the-art alignment methods. Additionally, OSP language models are sample-efficient, generalize effectively to OOD tasks, and have the potential for self-improvement.
\end{itemize}

% Then we construct response pairs in a ranked pairing manner.

% whose objective is to minimize the error between implicit preference strength predicted by the current policies and self-preferring strength.

% language models achieve state-of-the-art alignment performances in both tasks and show greater advantages when the amount of data is smaller ($2\%$).

% Besides, OSP language models can efficiently self-improve on downstream tasks without external supervision when established with LLMs proficiency in self-preferring.

% , which can mitigate overfitting and improve performance.

\section{Preliminaries}
In this section, we summarize the key processes of two representative LLM alignment methods: RLHF and DPO. 
Given a prompt $\bx \in \mathcal{X}$, there are corresponding highly preferred and less preferred responses $\by^h, \by^l \in \mathcal{Y}$ in the human preference dataset $\dataset$, where $\mathcal{X}$ and $\mathcal{Y}$ are the finite spaces of prompts and responses, respectively. 
Mathematically, based on $\dataset$, alignment methods aim to train the LLM $\pi_{\theta}$: $\mathcal{X} \rightarrow \mathcal{Y}$ that align with human preferences. 
Following ~\cite{IPO}, the ground-truth preference strength for any two responses $\by, \by' \in \mathcal{Y}$ is defined as $p^*(\by \succ \by' | \bx) = \E_j [{\mathbb{I}\{j \ \text{prefers} \ \by \ \text{to} \ \by' \  \text{given} \  \bx}\}]$, where $\mathbb{I}$ is the indicator function, and the expectation is over humans $j$.
Note that preference strength information $p^*(\by \succ \by' | \bx)$ is commonly not available in $\dataset$ and existing methods only utilize binary preferences instead.

% and model the preference strength of $\by^h$ over $\by^l$ as $\by^h \succ \by^l$.
% binary preferences instead of $p^*(\by \succ \by' | \bx)$ are commonly available in $\dataset$.
% and the preference strength information $p^*(\by \succ \by' | \bx)$ is not used in existing methods. 

% We consider the alignment problem for LLM parameterized by $\theta$ and denoted by $\pi_{\theta}$.
% In the offline human preference dataset $\dataset$, given prompt $\bx \in \mathcal{X}$, there are corresponding human label highly preferred and less preferred responses $\by^h, \by^l \in \mathcal{Y}$, where $\mathcal{X}$ and $\mathcal{Y}$ are the finite spaces of prompts and responses, respectively.
% We denote LLM after supervised fine-tuning SFT as $\pi_{\text{SFT}}$.
% Following ~\cite{IPO}, the ground-truth preference strength probabilities can be defined as $p^*(\by \succ \by' | \bx) = \E_j [{\mathbb{I}\{j \ \text{prefers} \ \by \ \text{to} \ \by' \  \text{given} \  \bx}\}]$, where $\mathbb{I}$ is the indicator function and the expectation is over humans $j$.
% We define the preference function $p: \mathcal{Y} \times \mathcal{Y} \rightarrow  [0,1]$ to represent the preference strength probabilities between response pairs.
% We proceed with the assumption that preference functions adhere to the symmetry condition $p(\by, \by') = 1-p(\by', \by) $.

\paragraph{Reinforcement Learning from Human feedback.}
RLHF~\cite{InstructGPT} introduces the three-step pipeline comprising supervised fine-tuning (SFT), reward modeling, and reinforcement learning (RL) fine-tuning with proximal policy optimization (PPO)~\cite{PPO}.
As illustrated in \cref{fig:comparison}, RLHF learns a substituted reward based on $\dataset$ to approximate the ground-truth preference strength.
When the Bradley-Terry model~\cite{bradley1952rank} holds for $p^*$, there exists a reward function $r: \mathcal{X} \times \mathcal{Y} \rightarrow  \mathbb{R}$ such that:
\begin{equation}
    p^*(\by \succ \by' | \bx) = \sigma(r(\bx, \by) - r(\bx, \by')),
    \label{equation:BT}
\end{equation}
where $\sigma$ is the sigmoid function.
The objective of RL fine-tuning is then to maximize the substituted rewards while maintaining the Kullback–Leibler (KL) distance between the current model $\pi_{\theta}$ and the reference model, usually served by SFT LLM $\pi_{\text{SFT}}$:

\begin{equation}
\mathop{\mathrm{arg \ max}}_{\pi_\theta} \E_{\bx \sim \promptdataset}\E_{\by \sim \pi_\theta(\cdot|\bx)} \big[r(\bx, \by) \big]  - \beta \KL (\pi_\theta|| \pi_{\text{SFT}}).
\label{equation:RLHF-objective}
\end{equation}

However, a separate reward model introduces parameter inefficiency, and RL objectives usually result in unstable training~\cite{Moss}.
Moreover, reward models may fail to express complex preference relations, such as non-transitive or cyclic preference~\cite{bertrand2023limitations,NLHF}.

\paragraph{Direct Preference Optimization.}
% The optimal solution $\pi^*$ to Equation~\ref{equation:RLHF-objective} is available in a closed form:
% \begin{equation}
%     \pi^*(\by|\bx)  \propto  \pi_{\text{SFT}}(\by|\bx) \cdot \mathrm{exp}\left(\frac{1}{\beta} r(\bx, \by) \right).
%     \label{equation:RLHF-optimal}
% \end{equation}

% Applying the expression of $r(\bx, \by)$ in Equation~\ref{equation:RLHF-optimal} to Equation~\ref{equation:BT},
% Applying the expression of $r(\bx, \by)$ in Equation~\ref{equation:RLHF-optimal} to Equation~\ref{equation:BT}, 
An alternative approach to RLHF described above is DPO, which bypasses the reward function and directly optimizes the objective in Equation~\ref{equation:RLHF-objective} by solving a binary classification problem on the offline human preference dataset $\dataset$.
\begin{equation}
\mathcal{L}_{\text{DPO}} (\pi_{\theta} ; \pi_{\text{SFT}}) = -\E_{\bx, \by^h, \by^l  \sim \dataset} \big[ \log \sigma \big( \beta \log(\frac{\pi_\theta(\by^h | \bx)}{\pi_{\text{SFT}}(\by^h | \bx)}) - \beta \log(\frac{\pi_\theta(\by^l | \bx)}{\pi_{\text{SFT}}(\by^l | \bx)}) \big) \big],
\end{equation}
where $\pi_{\text{SFT}}$ is used to normalize the logits.
DPO and its variants overlook the preference strength information within the response pairs.
% While practical for commonly collected offline human preference datasets.
Besides, offline methods do not utilize on-the-fly-generated samples to improve data coverage and quality, resulting in weaker peak performance compared to online methods~\cite{tang2024understanding}.

\section{Method}

In this section, we detail our proposed online self-preferring language models, which utilize the self-judged preference strength information of self-generated response pairs.
% In this section, we detail our proposed online self-preferring language models, which utilize LLMs' self-judge capability to capture soft preference strength information from self-generated preference pairs.
We first introduce how our method models preference strength, deducing the soft-preference cross-entropy loss function.
Then, we illustrate the overall framework and core designs of our methods.
% We hypothesize that such information is critical for avoiding overfitting and improving alignment performance.
% As shown in~\cref{fig:arch}, our core designs include on-policy response sampling, ranked pairing response pair construction, self-preferring, and preference-strength aware loss function.

% In this section, we first derive the soft-preference cross-entropy loss based on the error between implicit preference strength predicted by the current policies and
% target preferring strength.
% Then we introduce the framework and key components of online self-preferring language models.

\subsection{Preference Strength Modeling}
\label{sec:method:spce}
Different from the RLHF objective described in Equation~\ref{equation:RLHF-objective}, our goal is to explicitly model the general preference instead of substituting preferences with rewards. The objective is as follows:
\begin{equation}
\mathop{\mathrm{arg \ max}}_{\pi_\theta} \E_{\bx \sim \promptdataset}\E_{\by \sim \pi_\theta(\cdot|\bx), \by' \sim \pi_\mu(\cdot|\bx)} \big[p^*(\by \succ \by' |\bx) \big]  - \beta \KL (\pi_\theta|| \pi_{\text{SFT}}),
\label{equation:OSP-objective}
\end{equation}
where $\pi_\mu$ denotes the LLM that samples responses.
% To optimize the OSP objective in Equation~\ref{equation:OSP-objective}, we need to specify the behavior policy to sample responses.
% Inspired by the recent self-play theories~\cite{SPIN, swamy2024minimaximalist} where one improves its policy by playing against oneself, we utilize the current policy as the behavior policy.
% \begin{equation}
% \mathop{\mathrm{arg \ max}}_{\pi_\theta} \E_{\bx \sim \promptdataset}\E_{\by \sim \pi_\theta(\cdot|\bx)}  \big[ \E_{\by' \sim SG(\pi_{\theta}(\cdot|\bx))} p^*(\by \succ \by' |\bx) \big]  - \beta \KL (\pi_\theta|| \pi_{\text{SFT}}),
% \label{equation:OSP-self-objective}
% \end{equation}
% where we placing a $stop\_gradient (SG)$ on sampling of the adversarial responses.
The closed-form solution of the KL-constrained objective in Equation~\ref{equation:OSP-objective} is:
\begin{equation}
    \pi^*(\by|\bx)  \propto  \pi_{\text{SFT}}(\by|\bx) \cdot \mathrm{exp}\left(\frac{1}{\beta} p^*(\by \succ \pi_{\mu} |\bx) \right),
    \label{equation:OSP-optimal}
\end{equation}
where we define $p^*(\by \succ \pi_{\mu} |\bx) := \E_{\by' \sim \pi_{\mu}(\cdot|\bx))} p^*(\by \succ \by' |\bx) $.
For any response pairs $\by^h \succ \by^l$, the optimal policy $\pi^*(\by|\bx)$ satisfies :
\begin{equation}
    \sigma \big( \beta \log(\frac{\pi^*(\by^h | \bx)}{\pi_{\text{SFT}}(\by^h | \bx)}) - \log(\frac{\pi^*(\by^l | \bx)}{\pi_{\text{SFT}}(\by^l | \bx)}) \big) = \sigma \big( p^*(\by^h \succ \pi_{\mu} |\bx) - p^*(\by^l \succ \pi_{\mu} |\bx) \big),
    \label{equation:OSP-two-distributions}
\end{equation}
where $\sigma: \mathbb{R} \rightarrow  [0,1]$ is the logistic function.
We denote the left-hand side in Equation~\ref{equation:OSP-two-distributions} as $\hat p (\by^h \succ \by^l | \bx, \pi^*)$, which represents the implicit preference strength predicted by the optimal policy $\pi^*$ and normalized by the reference policy $\pi_{\text{SFT}}$.
On the right-hand side, we define the target preference strength as $p(\by^h \succ \by^l |\bx)$.
% \begin{equation}
%     p(\by^h \succ \by^l |\bx):=\sigma \big( p^*(\by^h \succ \pi_{\mu} |\bx) - p^*(\by^l \succ \pi_{\mu} |\bx) \big).
% \end{equation}
Both implicit and target preference strengths are probabilities.

\paragraph{Soft-Preference Cross-Entropy Loss.}
% \label{sec:spce-loss}
In practice, the optimal policy $\pi^*$ is not available.
Without $\hat p (\by^h \succ \by^l | \bx, \pi^*)$, the implicit preference strength can be predicted with the current policy $\pi_\theta$:
\begin{equation}
\begin{aligned}
\label{eq:predicted-ps}
    \hat p (\by^h \succ \by^l | \bx, \pi_\theta)  =  \sigma \big( \beta \log(\frac{\pi_\theta(\by^h | \bx)}{\pi_{\text{SFT}}(\by^h | \bx)}) - \log(\frac{\pi_\theta(\by^l | \bx)}{\pi_{\text{SFT}}(\by^l | \bx)}) \big),
\end{aligned}
\end{equation}
To derive the SPCE loss function, we take the KL-divergence $\KL (p(\by^h \succ \by^l | \bx) || \hat p(\by^h \succ \by^l | \bx, \pi_\theta))$ between the target and implicit preference strength. This results in:
\begin{equation}\label{eq:OSP-loss}
\begin{aligned}
    \mathcal{L}_\text{SPCE}(\pi_{\theta})  = \E_{\by^h, \by^l \sim \pi_{\theta}}  [  p(\by^h \succ & \by^l )  \log \big( \hat p (\by^h \succ \by^l | \pi_\theta) \big) + \\ & (1-p(\by^h \succ \by^l)) \log \big(1- \hat p (\by^h \succ \by^l | \pi_\theta) \big) ],
\end{aligned}
\end{equation}
where we suppress the dependence on $\bx$ in preference strength to denote the expectation of the prompt distribution.
In Equation~\ref{eq:OSP-loss}, we employ the response pairs sampled from the current model $\pi_\theta$,  specifically utilizing on-policy sampling, for computing the SPCE loss. 
During this process, we apply a stop gradient to the sampling.
%, where we place a stop gradient on the sampling.
% Such an on-policy manner offers the potential for exploration~\cite{lambert2022challenges}.
% We construct multiple response pairs for an emerging prompt $\bx$. 
% Besides, utilizing samples generated by the model itself .

% Similar to \cite{IPO}, we can demonstrate that optimizing $\pi_{\theta}$ using SPCE loss converges to $\pi^*$.
% \begin{theorem}
% \label{theorem:spce-loss}
%     Define $\Pi$ to be the set of policies $\pi$ such that $\texttt{Supp}(\pi_{\theta})= \texttt{Supp}(\pi_\text{SFT})$, then $\pi_{\theta} \mapsto \mathcal{L}_\text{SPCE}(\pi_{\theta})$ has a unique local/global minimum in $\Pi$, which is $\pi^*$. 
% \end{theorem}
% We refer to \cref{proof:spce-loss} for a detailed discussion and proof.

\textbf{Gradient Analysis.}  
To analyze what our proposed SPCE loss update does, we compare its gradient with respect to the parameters $\theta$ to other pair-wise alignment methods.
The gradient of our SPCE loss is as follows:
\begin{equation}\label{eq:OSP-gradient}
\begin{aligned}
    \nabla_\theta \mathcal{L}_\text{SPCE}(\pi_{\theta}) = \E_{\by^h, \by^l \sim \pi_\theta} [\big( \hat p (\by^h \succ \by^l | \pi_\theta) )  - p(\by^h \succ \by^l) \big)  \big( \nabla_\theta \log \pi_\theta(\by^h  )- \nabla_\theta \log \pi_\theta(\by^l  ) \big)].
\end{aligned}
\end{equation}
The gradient of the SPCE loss becomes zero when $\hat p (\by^h \succ \by^l | \pi_\theta) ) = p(\by^h \succ \by^l) $, indicating that the model's predicted preference strength matches the target self-preference strength.
The SPCE loss is utilized to train the model whenever there is a discrepancy between them, enhancing the model's awareness of preference strength.
% is aware of the preference strength, which trains the policy if there is an error between the predicted and the target preference strength.
In comparison, the gradient of DPO is as follows:
\begin{equation}\label{eq:DPO-gradient}
\begin{aligned}
    \nabla_\theta \mathcal{L}_\text{DPO}(\pi_{\theta}) = \E_{\by^h, \by^l \sim \dataset}  [\big( 1- \hat p (\by^h \succ \by^l | \pi_\theta) \big)  \big( \nabla_\theta \log \pi_\theta(\by^h )- \nabla_\theta \log \pi_\theta(\by^l ) \big) ].
\end{aligned}
\end{equation}

DPO trains the model until $\hat p (\by^h \succ \by^l | \bx, \pi_\theta)$ reaches the maximum value $1$, which necessitates $\log(\frac{\pi_\theta(\by^h | \bx)}{\pi_\theta(\by^l | \bx)})  \rightarrow \infty$.
Consequently, DPO is susceptible to overfitting. To mitigate this drawback, IPO~\cite{IPO} introduces a hard margin to enforce the gradient to approach zero when the preference strength reaches a certain threshold:
% \begin{equation}\label{eq:CDPO-gradient}
% \begin{aligned}
%     \nabla_\theta \mathcal{L}_\text{IPO}(\pi_{\theta}) =  \E_{\by^h, \by^l \sim \dataset}  [\big(\log(\frac{\pi_\theta(\by^h )}{\pi_{\text{SFT}}(\by^h )}) - & \log(\frac{\pi_\theta(\by^l )}{\pi_{\text{SFT}}(\by^l)}) -\frac{1}{2\beta} \big) \\ &
%     \big( \nabla_\theta \log \pi_\theta(\by^h )- \nabla_\theta \log \pi_\theta(\by^l ) \big)].
% \end{aligned}
% \end{equation}
\begin{equation}\label{eq:IPO-gradient}
\begin{aligned}
    \nabla_\theta \mathcal{L}_\text{IPO}(\pi_{\theta}) =  \E_{\by^h, \by^l \sim \dataset}  [\big( \sigma^{-1}(\hat p (\by^h \succ \by^l | \pi_\theta)) -\frac{1}{2\beta} \big) 
    \big( \nabla_\theta \log \pi_\theta(\by^h )- \nabla_\theta \log \pi_\theta(\by^l ) \big)].
\end{aligned}
\end{equation}
Both DPO and IPO utilize binary preferences. In contrast, the soft preference employed by the SPCE loss utilizes a dynamic margin to regulate the updates for different response pairs.

% \begin{equation}\label{eq:CDPO-gradient}
% \begin{aligned}
%     \nabla_\theta \mathcal{L}_\text{RLHF}(\pi_{\theta}) =  \E_{\by^h, \by^l \sim \dataset} .
% \end{aligned}
% \end{equation}

\begin{figure*}[t]
    % \vskip 0.2in
    \begin{center}
\centerline{\includegraphics[width=0.92\columnwidth]{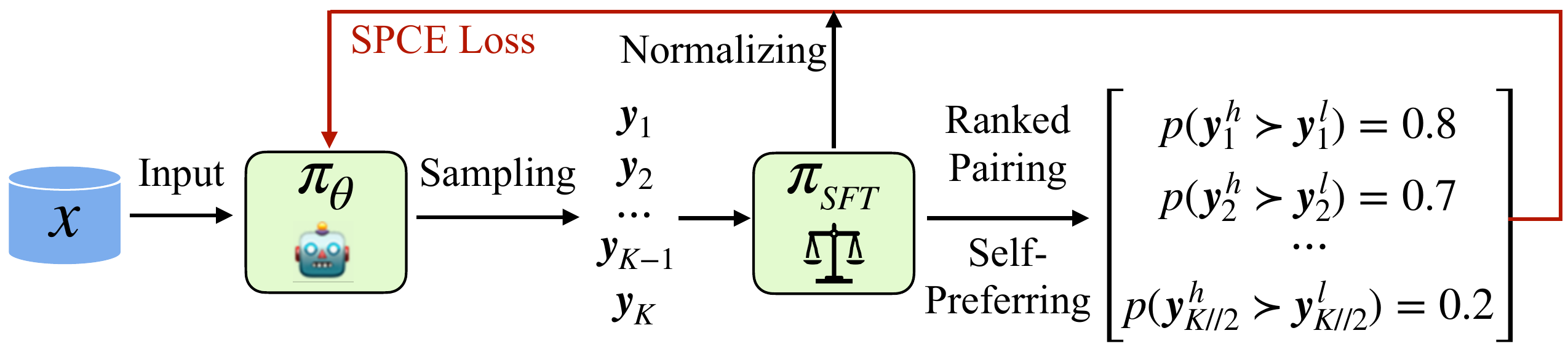}}
    \vspace{-0.5em}
    \caption{Illustration of the aligning pipeline of OSP. For each training prompt $\bx$, OSP first employs the current model $\pi_\theta$ to sample $K$ candidate responses, constructing $K//2$ response pairs in a ranked pairing manner.
    $\pi_\text{SFT}$ subsequently judges response pairs to obtain the preference strength.
    % , constructing $K//2$ response pairs with different preference strengths.
    Finally, our proposed SPCE loss leverages multiple response pairs with different preference strengths to align LLM $\pi_\theta$, where $\pi_\text{SFT}$ is also used for normalization.
    }
    \label{fig:arch}
    \end{center}
    \vspace{-1em}
\end{figure*}

\subsection{Online Self-Preferring Language Models}
OSP language models trained with the SPCE loss can effectively learn from response pairs exhibiting varying preference strengths.
As shown in~\cref{fig:arch}, we employ the on-policy manner to sample $K$ responses for each prompt using the current LLM $\pi_\theta$, aiming to enhance the prompt efficiency.
We provide a detailed explanation of our methodology for constructing pairs and determining target preference strengths below.

\paragraph{Ranked pairing.}
In accordance with the SPCE loss defined in \cref{eq:OSP-loss}, constructing response pairs that cover as many preference strength probabilities as possible benefits the prompt efficiency.
However, conducting comparisons among all the on-the-fly self-generated response pairs results in $\binom{K}{2}$ forward passes for comparisons. 
To reduce the computation overhead, we propose a ranked pairing method.
Specifically, we first evaluate the preference strength of each response in comparison to the blank baseline response, denoted as $p(\by \succ \varnothing |\bx)$.
Subsequently, we rank the $K$ responses based on their $p(\by \succ \varnothing |\bx)$ values, pairing the responses with the highest and lowest values as $\by_1^h, \by_1^l$, the second highest and lowest values as $\by_2^h, \by_2^l$, and so forth.
This approach allows us to efficiently generate $K//2$ response pairs that encompass a broad range of preference strengths, leading to a computational complexity reduction to $\mathcal{O}(K)$. 
% Furthermore, our empirical demonstration showcases that the ranked pairing method enhances both prompt efficiency and alignment performance.

% rank the responses according to their preference strengths compared to the blank baseline response

% Then we combine the highest and least preferred responses as $\by_1^h, \by_1^l$, the second highest and the least preferred responses as $\by_2^h, \by_2^l$, and so on.
% In this way, we can efficiently construct $K//2$ response pairs that cover preference strength as many as possible, which is critical for improving prompt efficiency and alignment performance.

% \paragraph{Response pairs construction.}
% OSP language models with the SPCE loss can learn from response pairs with various preference strengths, 
% so OSP generates $K$ responses for each prompt to construct response pairs.
% This can improve prompt efficiency, which means that LLMs can be aligned with a small number of prompts.  

% % ~\footnote{Similar to the reward model usage in \url{huggingface.co/stanfordnlp/SteamSHP-flan-t5-xl}}
% Then we adopt a ranked pairing manner for response pairs construction, i.e., combining the highest and least preferred as $\by_1^h, \by_1^l$, the second highest and the least $\by_2^h, \by_2^l$, and so on.
% In this way, we can efficiently construct $K//2$ response pairs that cover preference strength as many as possible.

\paragraph{Self-preferring.}
Recall that we define the target preference strength as follows:
\begin{equation}
    p(\by^h \succ \by^l |\bx):=\sigma \big( p^*(\by^h \succ \pi_{\mu} |\bx) - p^*(\by^l \succ \pi_{\mu} |\bx) \big).
\end{equation}
It's impractical to have access to the precise value of $p^*(\by \succ \pi_{\mu} |\bx)$ across the entire $\pi_{\mu}$ distribution.
Therefore, we opt to directly estimate the target preference strength $p(\by^h \succ \by^l |\bx)$ using a self-preferring manner with LLM-as-a-judge instructions.
To achieve this, we input the constructed instruction shown in~\cref{fig:hh-evalprompt} to LLM and extract the logit probabilities of generating the tokens of ``A'' or ``B'', followed by a softmax computation to derive the target preference strength $p(\by^h \succ \by^l |\bx)$.
$p(\by^h \succ \by^l |\bx)$ does not hinge on the Bradley-Terry model assumption, thus potentially capturing a wider array of human preference relations, including non-transitive or cyclic preferences.
% Compared to the Bradley-Terry reward model, $p(\by^h \succ \by^l |\bx)$ is not guaranteed to be transitive and can model complex preference relations such as non-transitive or cyclic preference.
% Then we can fine-tune the underlying LLM using SPCE loss proposed in Section~\ref{sec:spce-loss}.

% \textbf{Preference augmentation.}
% We adopt the LLM-as-a-judge manner for self-preferring.
% As shown in ~\cref{fig:hh-evalprompt} and \cref{fig:TLDR-evalprompt}, we state the principles that represent human preferences and construct the instruction by filling in the placeholders highlighted in red.
Without instruction tuning, most LLMs that are not sufficiently large cannot effectively follow the LLM-as-a-judge instructions, whose judging performance is not significantly better than random guessing.
To overcome this limitation, we create a seed set of LLM-as-a-judge instructions using an offline human preference dataset.
We then enhance the preference capabilities of LLMs using these instructions, finding that even a small amount of data can greatly enhance the LLM-as-a-judge ability.
It should be noted that preference augmentation can be omitted for LLMs proficient in preference selection.
Algorithm~\ref{alg:osp} outlines the general framework of online self-preference language models.

\begin{algorithm}[t]
	\caption{Framework of Online Self-Preferring Language Models\label{alg:osp}}
	\begin{algorithmic}[1]
		\STATE {\bfseries Require:} Prompt dataset $\dataset$, SFT model $\pi_{\text{SFT}}$, number of training steps $T$.
    \STATE Initialize $\pi_\theta=\pi_{\text{SFT}}$
    \STATE Augment LLM-as-a-judge ability for $\pi_{\text{SFT}}$ (Optional): supervised fine-tuning $\pi_{\text{SFT}}$ with LLM-as-a-judge instructions constructed by human preference seed dataset.
    \FOR{training step $1, \dots, T$}
    \STATE Select a prompt $\bx$ from $\dataset$.
    \STATE Sample $K$ responses using $\pi_\theta$: $\by_1, \cdots, \by_K \sim \pi_{\theta}(\by|\bx)$.
    \STATE Rank the responses according to $p(\by \succ \varnothing |\bx)$ and construct $K//2$ pairs.
    % \STATE Construct $K//2$ response pairs with the ranked pairing method.
    \STATE Preferring the target preference strength of the response pairs using $\pi_{\text{SFT}}$: $p(\boldsymbol{y}^h \succ \boldsymbol{y}^l  |\bx)$.
    \STATE Calculate the predicted preference strength $\hat p (\by^h \succ \by^l | \bx, \pi_\theta)$ using \cref{eq:predicted-ps}.
    \STATE Update $\pi_\theta$ using $\nabla_\theta \mathcal{L}_\text{SPCE}(\pi_\theta)$ in Equation~\ref{eq:OSP-gradient}.
    % \STATE Update $\pi_\theta$ using $\nabla_\theta \mathcal{L}_\text{PS}(\bz, \by')$
    \ENDFOR
    \STATE {\bfseries Return:} Aligned OSP language model $\pi_\theta$.
	\end{algorithmic}
\end{algorithm}

\section{Experiments}
\label{sec:exp}
In this section, we empirically validate the efficiency of OSP through extensive experiments conducted on two popular NLP tasks, namely Anthropic helpfulness and OpenAI summarization.

\subsection{Experimental Setups}
% For the helpful dialogue task, models of all methods are established using pre-trained Tiny-Llama~\footnote{\url{https://huggingface.co/TinyLlama/TinyLlama-1.1B-intermediate-step-1431k-3T}}.
% In order to enable Tiny-Llama to have self-feedback capabilities, we construct instructions with the format in Figure~\ref{fig:hh-evalprompt} based on 40\% pairwise samples in the Anthropic Helpful dataset~\cite{bai2022training}.
% We then conduct instruction tuning for Tiny-Llama before self-improving.

% For the summarization task that requires a higher context length, we use the Reddit ``TL;DR'' (Too Long; Didn't Read) dataset~\cite{TLDR}.
% We establish the model using instruction fine-tuned Mistral-7B~\footnote{\url{https://huggingface.co/mistralai/Mistral-7B-Instruct-v0.2}} and do not conduct instruction tuning because we observe that Mistral-7B already has satisfactory self-feedback ability (ACC=63\% on the test set) on the ``TL;DR'' dataset.
We primarily evaluate OSP and baselines built on the popular LLM TinyLlama\footnote{\url{https://huggingface.co/TinyLlama/TinyLlama-1.1B-intermediate-step-1431k-3T}}.
Additionally, we extend Mistral-7B-Instruct~\footnote{\url{https://huggingface.co/mistralai/Mistral-7B-Instruct-v0.2}} that is proficient in preferring.
We leverage two distinct datasets tailored for different tasks. 
For the helpful dialogue task, we employ the Anthropic Helpful (HH) dataset~\cite{bai2022training}. In this dataset, $\bx$ signifies a segment of conversation involving interactions between a human and a digital assistant. The goal is to train the model to generate a helpful response $\by$ in the subsequent turn.
For summarization, our experiment uses the Too Long; Didn't Read (TL;DR) dataset~\cite{TLDR} proposed by OpenAI. Here, $\bx$ represents a forum post, while $\by$ corresponds to its summary.
% In order to enable Tiny-Llama to have self-feedback capabilities, we construct instructions with the format in Figure~\ref{fig:hh-evalprompt} based on 40\% pairwise samples in the Anthropic Helpful dataset~\cite{bai2022training}.
Hyper-parameters and other experimental details are detailed in Appendix~\ref{appendix:detailed-exp-settings}.

\subsection{Alignment Performance Evaluations}
\label{sec:alignment-performance}
% In this section, we demonstrate that OSP is a sample-efficient and state-of-the-art alignment method, and has strong generalization abilities.
% For all the methods in \cref{tab:main:performance}, we start from the supervised fin-tuned models with $1\%$ preferred samples from the offline dataset. 
We compare OSP with state-of-the-art alignment baselines named SFT, DPO, IPO, and RLHF.
The SFT baseline utilizes prompts and exclusively human-preferred answers for supervised learning. 
DPO~\cite{DPO} employs the contrastive loss of pairwise responses and binary preference labels to align with human preferences.
IPO~\cite{IPO} incorporates a constant margin into the DPO loss function to prevent overfitting to offline datasets.
Unlike DPO and IPO directly fine-tuning LLMs with offline datasets, RLHF~\cite{InstructGPT} firstly learns a reward model and then uses the online RL algorithm PPO~\cite{PPO} to optimize LLMs.
It is important to note that RLHF requires an additional reward model to be maintained, while OSP maintains the same parameter efficiency as offline methods.

\begin{table}[t]
	\centering
 \caption{
 We compare the performance of OSP with baselines established by TinyLlama using model-based evaluation metrics. Specifically, we consider scores of SteamSHP-flan-t5-xl~\cite{pmlr-v162-ethayarajh22a} score for the helpfulness task and fine-tuned GPT-J-6B~\cite{fine-tuned-GPTJ} along with automatic metrics ROUGE for the summarization task. We highlight the best-performing values. All experiments are repeated four times with different seeds, and we report the average scores and standard deviations. \label{tab:main:performance}}
  \vskip -0.04in
  \renewcommand{\arraystretch}{1.1} % 增加行间距
   \resizebox{0.95\linewidth}{!}{
\begin{tabular}{l|c|c|c |c |c c}
\toprule
\multirow{2}{*}{Method}  & \multirow{2}{*}{Venue}  & Online  & Preference  & {HH}  & \multicolumn{2}{c}{TL;DR} \\
% \cmidrule(lr){6-7} \cmidrule(lr){8-9}
&&Method  & Type & Score   & Score & ROUGE  \\  \hline
SFT  &-&  No & N.A.       & $0.919$$\pm0.001$      &$1.93$$\pm0.01$ & $0.271$$\pm0.004$                                      \\
DPO   &NeurIPS 2023~\cite{DPO}& No &  Binary       & $0.896$$\pm0.041$ &  $2.24$$\pm0.09$ & $0.273$$\pm0.007$                                    \\
IPO    &arXiv 2023~\cite{IPO}& No &  Binary      & $0.925$$\pm0.004$ & $2.44$$\pm0.03$ & $\textbf{0.290}$$\pm0.012$  \\  
RLHF   &NeurIPS 2022~\cite{InstructGPT}& Yes & Reward  &  $0.932$$\pm0.007$   & $2.78$$\pm0.11$ & $0.210$$\pm0.003$                                      \\
% Self-Judge &arXiv 2024~\cite{Self-judge}& Yes  & Binary \\
OSP   &Ours& Yes  & Soft     &   $\textbf{0.958}$$\pm 0.006$   &  $\textbf{2.97}$$\pm0.06$   &  $0.275$$\pm0.007$                   \\ 
% OSP   &Ours, $2\%$ Dataset& Yes  & Soft     &   $\textbf{0.955}$$\pm 0.002$   &  $\textbf{3.01}$$\pm0.06$   &  $0.257$$\pm0.007$                   \\ 
\bottomrule
\end{tabular}
}
\vskip -0.1in
\end{table}

\paragraph{OSP is a strong alignment method.}
Model-based and automatic evaluation results are presented in \cref{tab:main:performance}.
It is observed that DPO experiences slight overfitting on the full dataset even after just one epoch, resulting in responses that are verbose and repetitive, as demonstrated in \cref{appendix:GPT4:OSP-DPO}.
Conversely, IPO avoids overfitting and consistently enhances the performance of DPO.
% on the full dataset, though the performance is slightly worse than DPO on the $2\%$ dataset.
RLHF achieves competitive performances compared to offline baselines except for the ROUGE metric.
This discrepancy arises because online methods do not directly fit the offline dataset and instead encourage exploration, whereas ROUGE measures the similarity between a generated summary and a reference summary from the dataset.
OSP language models significantly surpass both online and offline baselines across all model-based metrics and improve ROUGE scores over RLHF in the summarization task.

\begin{wrapfigure}{h}{0.61\textwidth}
	\centering
	\vspace{-0.5cm}
	\includegraphics[width=0.61\textwidth]{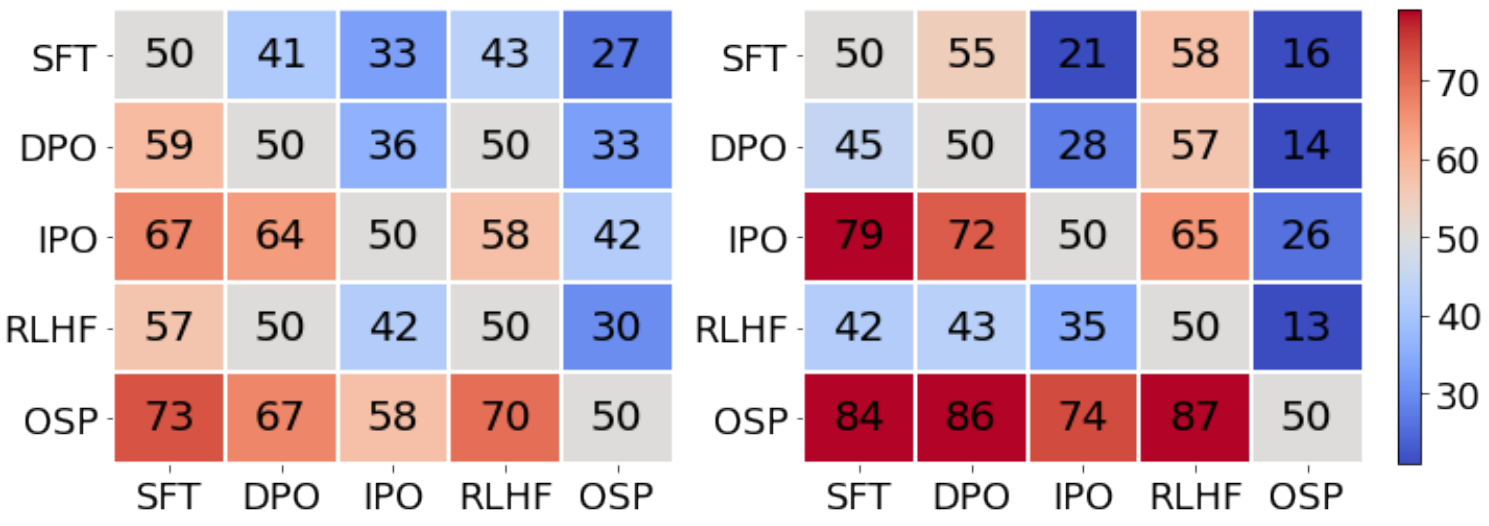}
	\vspace*{-0.4cm}
	\caption{\small Head-to-head comparison on HH (left) and TL;DR dataset (right), where win rates (\%) are evaluated by GPT-4.
    The winning rate indicates the percentage of samples of methods on the vertical axis that outperform those on the horizontal axis.
    \label{fig:Head2Head}}
	\vspace*{-0.4cm}
\end{wrapfigure}

To further assess alignment performance, we evaluate the responses of LLMs trained with different methods using GPT-4~\cite{GPT4}, which has been shown to be consistent with human preferences.
% The prompt for GPT-4 evaluation can be found in \cref{appendix:GPT-4-prompts}.
Results in \cref{fig:Head2Head} demonstrate that OSP is a strong alignment method, consistent with the conclusions drawn from model-based metrics.
Case studies are available in~\cref{appendix:GPT-4-evaluations-HH}.
Notably, although the ROUGE scores of summaries generated by OSP are lower than those of IPO, GPT-4 prefers 74 out of 100 OSP summaries over those from IPO.
This indicates that OSP language models generate higher-quality summaries than the offline dataset, highlighting the advantages of online methods.

\paragraph{OSP language models are sample efficient.}
Online methods can generate on-the-fly responses and preferences for alignment, reducing the need for extensive and high-quality data coverage.
We first ablate portions of the dataset to examine its impact on the accuracy of reward models and self-preference models.
Results in \cref{tab:RMEvaluation} show that preference models consistently outperform reward models across various portions of both datasets, suggesting that preference models with LLM-as-a-judge are more sample efficient than reward models.
% Besides, OOD generalization evaluations in \cref{sec:alignment-performance} reveal that preference models are more robust to avoid overoptimization.

\begin{wraptable}{r}{0.5\textwidth}
    \vspace{-7mm}
    \small
    \caption{Alignment performance with $2\%$ data. \label{table:002dataset}}
    % \vskip -0.07in
    % \vspace{-1mm}
    \renewcommand{\arraystretch}{1.1} % 增加行间距
    \resizebox{\linewidth}{!}{
    \begin{tabular}{l|c|cc}
        \toprule
        & HH & \multicolumn{2}{c}{TL;DR}  \\  %\cmidrule(lr){2-2}
        & Score & Score & ROUGE  \\ \hline %\cmidrule(lr){2-2} \cmidrule(lr){3-4}
        RLHF & $0.924 \pm 0.005$ & $1.82$$\pm0.50$  & $0.209$$\pm0.029$ \\
        OSP & $\textbf{0.955}$$\pm0.002$ &$\textbf{3.01}$$\pm0.19$   & $0.257$$\pm0.012$  \\ 
        \bottomrule
    \end{tabular}
    }
\end{wraptable}

We also evaluate the alignment models of RLHF and OSP when only a small portion of data ($2\%$) is available, corresponding to 1525 samples in the HH dataset and 1850 samples in the TL;DR dataset.
For preference augmentation, we utilize $1\%$ human preference samples to construct LLM-as-a-judge instructions, where response A is deemed better than response B, and utilize another $1\%$ by swapping the positions of responses.

Surprisingly, we do not observe a significant performance degradation when the amount of data is reduced to $2\%$, even when the accuracies of self-preferring decrease.
However, RLHF is ineffective with such a small dataset, especially in the summarization task.
This is because the accuracy of the reward model learned with $2\%$ TL;DR dataset is only $53\%$, which is insufficient for RL fine-tuning.

% Considering both RLHF and OSP start from the same SFT checkpoint, this is because the reward model trained with $2\%$ TL;DR dataset is not robust enough for RL fine-tuning.

% In addition to evaluating the full dataset, we randomly select $2\%$ offline human preference dataset for preference augmentation, corresponding to 1525 samples for the helpful dialogue task and 1850 samples for the summarization task.

\begin{table}[t]
    \centering
    \caption{Comparison of accuracy between preference and reward models with various dataset portions.\label{tab:RMEvaluation}}
    \vskip -0.07in
    \renewcommand{\arraystretch}{1.1} % 增加行间距
    \resizebox{0.95\linewidth}{!}{
    \begin{tabular}{l|cc|cc|cc|cc|cc}
        \toprule
        & \multicolumn{2}{c|}{$0\%$ Data}& \multicolumn{2}{c|}{$2\%$ Data} & \multicolumn{2}{c|}{$10\%$ Data} & \multicolumn{2}{c|}{$50\%$ Data} & \multicolumn{2}{c}{$100\%$ Data} \\ \hline
        % \cmidrule(lr){1-11} 
        % \cmidrule(lr){4-5}\cmidrule(lr){6-7}\cmidrule(lr){8-9}\cmidrule(lr){10-11}
        & HH & TL;DR & HH & TL;DR & HH & TL;DR & HH & TL;DR& HH & TL;DR\\ 
        Preference Model &$45\%$&$52\%$& $57\%$ &$62\%$ & $61\%$ &$67\%$  & $66\%$ & $71\%$ & $70\%$ & $71\%$ \\  % 20%TLDR上是69\%
        Reward Model &$41\%$&$44\%$& $53\%$ & $57\%$ & $56\%$ & $60\%$ &$62\%$ &$65\%$&$62\%$ & $70\%$ \\ 
        \bottomrule
    \end{tabular}
    }
\end{table}

\paragraph{OSP language models can effectively generalize to out-of-domain tasks.}
RLHF and OSP leverage the generalization abilities of LLMs in both generation and judgment, enhancing OOD tasks without relying on new human preferences.
The self-preferring and reward models are trained on the HH dataset.
We then evaluate their OOD generalization ability on two SHP datasets~\cite{SHP} from the domains named ``askbaking” and ``askacademia”.
Since SteamSHP-flan-t5-xl is fine-tuned on both HH and SHP datasets, we can still utilize it to proxy human preferences.
% For both methods, we utilize the $1\%$ HH dataset for SFT and another $2\%$ for learning a self-preference or separate reward model.
% ``askphilosophy”
% We regard two domains named ``askbaking” and ``askacademia” of the SHP dataset~\cite{SHP} as OOD tasks.

\begin{figure*}[t]
    \subfigure[Askbaking.
    % by reward models and gold reward models.
        \label{fig-askbaking}]
        {
            \centering
            \includegraphics[width=0.3\linewidth]{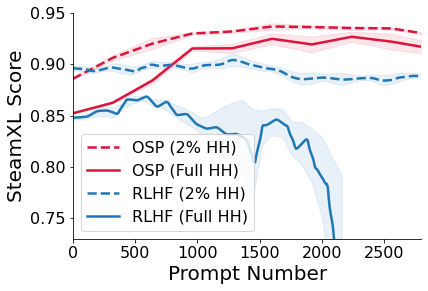}
        }
        \quad
    \subfigure[Askacademia.
    \label{fig-askacademia}]
    {
        \centering
        \includegraphics[width=0.3\linewidth]{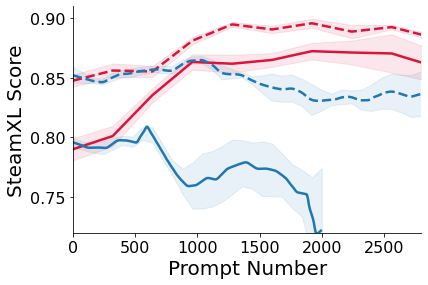}
    }
            % \quad
    % \subfigure[Askphilosophy.
    % \label{fig-askphilosophy}]
    % {
    %     \centering
    %     \includegraphics[width=0.23\linewidth]{figures/OOD/askphilosophy_with-full.png}
    % }
     \quad
    \subfigure[w/o preference augmentation.
    \label{fig-zero-shot}]
    {
        \centering
        \includegraphics[width=0.3\linewidth]{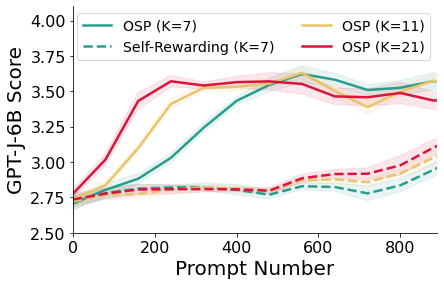}
    }
    \vspace{-1em}
    \caption{
   (a$\sim$b) Alignment of LLMs established by TinyLlama on OOD generalization tasks. (c)  Alignment of LLMs established by Mistral-7B-Instruct-v0.2 without using a human preference dataset.  
   All curves are averaged across 4 seeds, and the shaded area indicates the standard deviation.
    \label{fig-OOD}}
\end{figure*}

In \cref{fig-askbaking} and \cref{fig-askacademia}, methods with $2\%$ dataset achieve better generalization than ones with the full dataset.
Notably, RLHF using the reward model trained with the full HH dataset encounters serious reward hacking problems~\cite{gao2023scaling}, where optimizing an LLM to pursue maximal training rewards provided by the reward model results in high-reward yet low-quality LLMs that do not align with genuine human preferences. (please refer to~\cref{appendix:RLHF-overoptimization} for curves of training rewards).
In comparison, OSP language models do not suffer from overoptimization in these two OOD tasks, confirming the hypothesis that modeling general preferences instead of substituted rewards can potentially avoid reward hacking~\cite{amodei2016concrete}.
% It is also observed that the performance on OOD tasks could decrease in the cases of fine-tuning with more data.
% However, after aligning LLMs using our method, performances on the OOD tasks can be effectively improved.

\paragraph{OSP language models can self-improve efficiently without preference augmentation.}
\cref{tab:RMEvaluation} indicates that preference augmentation is crucial for TinyLlama.
Once the LLM reaches a certain level of proficiency, it becomes possible to bypass preference augmentation and proceed directly to self-preferring and self-improvement.
We verify this assumption by conducting experiments on summarization tasks with Mistral-7B-Instruct-v0.2, which achieves an accuracy of $67\%$ on the TL;DR test set without any external training data.

Both offline and online methods mentioned above, except for OSP, cannot perform alignment without high-quality responses or human preferences.
Therefore, we compare OSP with Self-Rewarding~\cite{yuan2024self}, where the LLM selects the highest and least preferred self-generated response to update LLM itself using DPO.
For a fair comparison, we re-implement Self-Rewarding in an online on-policy manner, using the same LLM-as-a-judge instruction as OSP.
As shown in \cref{fig-zero-shot}, OSP can significantly improve the summarization ability of fine-tuned Mistral-7B-Instruct-v0.2.
Additionally, OSP is more prompt-efficient than Self-Rewarding, especially when we sample more responses for each prompt, i.e., a larger sample number $K$.
We evaluate the win rate of the OSP language model using 200 prompts with $K=21$ against fine-tuned Mistral-7B-Instruct-v0.2 via GPT-4.
We observe significant self-improvement and the win rate is $95\%$.
Case studies can be found in Appendix~\cref{appendix:GPT-4-evaluations-TLDR}.
% Note that both offline and online methods mentioned above except for OSP can not perform alignment in this setting.

% Therefore, we only plot the training curve of OSP in \cref{fig-zero-shot}.

\subsection{Ablation Studies}

\begin{figure*}[t]
    \subfigure[Score of OSP with different loss functions.
    % by reward models and gold reward models.
        \label{fig-ablation-spce}]
        {
            \centering
            \includegraphics[width=0.3\linewidth]{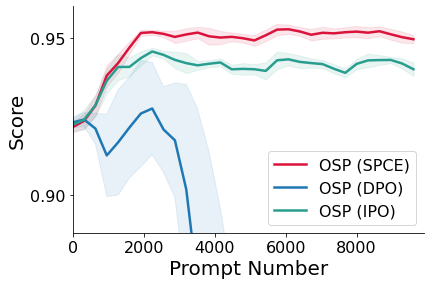}
        }
        \quad
    \subfigure[Score of OSP with different pair construction methods.
    \label{fig-ablation-pair-construction}]
    {
        \centering
        \includegraphics[width=0.3\linewidth]{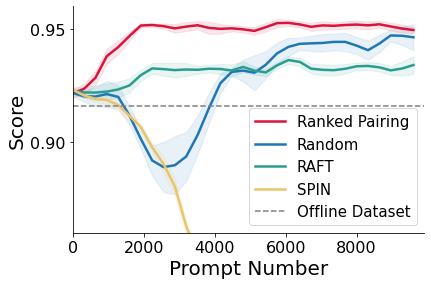}
    }
            \quad
    \subfigure[Score of OSP with different response sampling methods.
    \label{fig-ablation-sampling}]
    {
        \centering
        \includegraphics[width=0.3\linewidth]{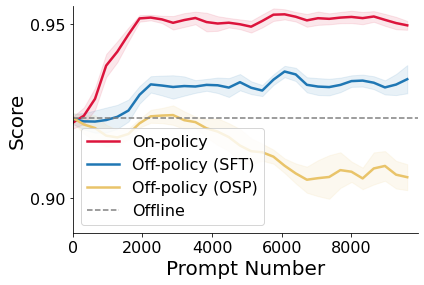}
    }
            % \quad
    % \subfigure[Sampling number.
    % \label{fig-zero-shot}
    % {
    %     \centering
    %     \includegraphics[width=0.23\linewidth]{figures/ablations/ablation-response-number.png}
    % }
    \vspace{-1em}
    \caption{
   Ablations of OSP language models established by TinyLlama trained on $2\%$ HH dataset with different loss functions, response sampling, and pair construction methods.
    \label{fig-step3-summarization-overoptimization}}
\end{figure*}

\paragraph{SPCE loss avoids overfitting and improves performance.}
To leverage the preference strength data embedded in self-assessed soft preference labels, we introduce the SPCE loss for OSP language models.
When translating soft preference labels into binary format, the SPCE loss reduces to the DPO loss, which is adopted by recent works such as OAIF~\cite{online-DPO} and SELF-JUDGE~\cite{Self-judge}.
We refer to this baseline as OSP (DPO).
As shown in \cref{fig-ablation-spce}, the DPO loss is not effective in the online self-preferring setting.
This indicates that the online setting is more vulnerable to overfitting (please refer to \cref{fig-loss} for loss curves).
Compared to DPO, IPO is more suitable for online learning.
Our proposed SPCE loss can bring further performance gains than the IPO loss due to utilizing the preference strength information.
% We also record the token length of OSP language models.
% It is observed that both the DPO loss make the response significantly longer than SPCE.
% Note that DPO is vulnerable to overfitting because of the Bradley-Terry modelization, which results in 
% Besides, we also observe that using more data for preference augmentation can 

\paragraph{Ranked pairing is effective and stable for training.}
We propose the ranked pairing pair construction method for OSP language models, which aims to cover as many preference strengths as possible.
We set a random baseline which pairs responses randomly.
RAFT~\cite{RAFT} collects multiple response pairs for each prompt and fine-tunes the LLM with the highest self-preferred response.
% (most preferred by LLM itself in our OSP language models).
% Max-Min collects multiple response pairs for each prompt, selects the highest and least preferred response to construct pairs, and then fine-tunes the LLM using SPCE loss.
SPIN~\cite{SPIN} requires additional SFT datasets and constructs pairs by always regarding self-generated responses as $\by^l$ and responses in SFT datasets as $\by^h$.

% \cref{fig-ablation-pair-construction} indicates that constructing response pairs with our proposed ranked pairing manner is most effective and stable for training.
\cref{fig-ablation-pair-construction} shows that random pairing can work because SPCE loss is aware of complementary preference strength, i.e., $p(\by \succ \by' |\bx) = 1-p(\by' \succ \by |\bx)$.
Our proposed ranked pairing manner is more effective and stable for training due to its broader coverage of preference strength probabilities.
RAFT converges to a lower performance because it does not utilize the less preferred responses.
We note that the performance of the SFT model surpasses that of the SFT dataset itself. Since a drawback of SPIN is its tendency to converge toward the distribution of the SFT dataset, constructing response pairs in SPIN's manner is ineffective in this setting.
% Despite achieving a competitive score with the ranked pairing method after being trained on 5000 prompts, Max-Min exhibits signs of overfitting during its initial training phases.

\paragraph{Effectiveness of online on-policy response sampling.}
OSP language models adopt the on-policy response sampling.
% , where the learning algorithm aligns LLMs based on response pairs generated directly by the LLM.
We also compare this on-policy approach with various sampling methods.
SFT sampling involves using samples exclusively from the SFT model for updating, which is adopted by SLIC-HF~\cite{zhao2023sliHF}.
RSO~\cite{RSO} samples responses via the optimal policy, achievable through rejected sampling from SFT sampling.
Additionally, we establish an offline baseline, which assigns soft preferences to response pairs in the offline dataset and utilizes the SPCE loss to align the LLM.

From \cref{fig-ablation-sampling}, it is observed that on-policy sampling is more effective than SFT sampling.
We hypothesize that exploring broader data coverage benefits online SPCE loss. 
On-policy sampling promotes exploration compared to SFT sampling and offline datasets.
Furthermore, after rejected sampling, RSO's exploration is further constrained, rendering it ineffective.

\section{Related Work}
\subsection{LLMs Alignment Methods}
% RLHF~\cite{InstructGPT} is a pivotal approach for fine-tuning language models to align with human preferences. 
% Inherited from the RL paradigm~\cite{levine2020offline}, 

Based on whether on-the-fly responses and feedback are collected, LLM alignment methods can be classified into online and offline categories.
The current dominant approach, RLHF~\cite{InstructGPT} is categorized under online training.
Many explorations have been undertaken on this path, aiming to improve training algorithm~\cite{NLPO,zhu2023fine,P3O}, reduce computational overhead~\cite{ReMax}, learn robust reward models~\cite{UP-RLHF} to mitigate overoptimization~\cite{gao2023scaling}.
Offline methods \cite{DPO, IPO, zhao2023sliHF, ethayarajh2024kto} emerged as an efficient alternative to RLHF, bypassing the reward modeling and directly fitting response pairs with binary preferences.
% , where a response is preferred or less preferred than the other.
% Notably, DPO~\cite{DPO} directly optimizes the LLMs toward the RLHF objective by using the contrastive loss of pairwise responses and binary preference labels in the offline dataset.

% Direct alignment methods aim to align LLMs from feedback with non-RL approaches to bypass the reward modeling step.
% Direct alignment methods can be broadly categorized into offline and online methods, according to whether on-the-fly samples are synthesized or not.
% Offline methods, such as DPO~\cite{DPO}, directly optimize LLMs towards the RLHF objective using contrastive loss.
% To mitigate the overfitting issue of DPO, IPO~\cite{IPO} adds a margin to the constrastive loss function.

Several studies apply the loss functions of offline methods to the online setting, differing in how to sample responses and construct response pairs.
With the help of reward models, RAFT~\cite{RAFT} samples multiple responses from the SFT model and selects the responses with the highest rewards to align LLMs using SFT, while RSO~\cite{RSO} proposes to sample responses from the optimal policies to construct response pairs and adopt either DPO or SLiC loss~\cite{zhao2023sliHF}.
SPIN~\cite{SPIN} does not rely on reward models, treating responses in SFT datasets as preferred and self-generated responses as less preferred to align LLMs using DPO loss.
Online on-policy response sampling has been demonstrated to be more effective for aligning LLMs~\cite{NLHF, DNO, online-DPO, tang2024understanding}.
We adopt the online on-policy manner and propose a ranked pairing method according to preference strengths to construct response pairs.

% Recent works demonstrate the effectiveness of the online manner for direct alignment methods.

% SLiC-HF~\cite{zhao2023sliHF} samples response pairs with the SFT model, then utilizes an external reward model to annotate the pairs.
% RSO~\cite{RSO} proposes to sample responses via the optimal policy estimated by the rejected sampling of SFT models and adopt either DPO or SLiC loss~\cite{zhao2023sliHF}.

% which is effective and efficient.
% However, online learning 
% We propose a new loss function, which is effective for the online on-policy setting and mitigating overfitting.
% However, these methods focus on playing against itself in the previous round to identify the Nash equilibrium policy, assuming a gold preference annotator is available.
% Our work also adopts the online on-policy manner, where soft preference strengths are provided for on-the-fly samples generated by the current policy.

% is an iterative DPO-like framework where human labels determine the high-quality response in a pair, while the generations from the last iteration serve as the low-quality response.

\subsection{Alignment from AI Feedback}
Given the high cost of human feedback, one promising strategy involves leveraging feedback from LLMs for supervision through LLM-as-a-judge instructions~\cite{zheng2023judging}.
RLCAI~\cite{bai2022constitutional} substitutes human feedback with feedback from a proficiency LLM conditioned solely on a set of written principles. 
RLAIF~\cite{lee2023rlaif} also leverages ``off-the-shelf'' LLMs to construct training datasets, and then follow the three-step RLHF pipeline to fine-tune LLMs.
OAIF~\cite{online-DPO} applies DPO to the online on-policy scenario, where online AI feedback is sourced from a more robust LLM annotator.
 
Recognizing that external LLMs are not always accessible, recent studies demonstrate the potential of utilizing feedback from the training LLM itself. 
RLC~\cite{pang2023language} employs self-evaluation feedback as rewards and optimizes LLMs with RL.
In the context of alignment, self-rewarding language models~\cite{yuan2024self} provide rewards for the generated offline dataset and then optimize LLMs with iterative DPO~\cite{iteratedDPO}. 
More recently, Self-JUDGE~\cite{Self-judge} fine-tuning LLMs to augment the judging ability to judge self-generated responses and also utilize DPO for optimization.
However, all methods mentioned above do not utilize the preference strength information that can be provided via LLM-as-a-judge.

% According to the rewards, preference pairs are used for training via DPO.

% We also utilize the feedback from the LLM itself
% Different from the above methods, we take advantage of the soft preferences.

% Feedback provided by LLM-as-a-judge can be directly used to fine-tune LLMs, using either RL manner~\cite{pang2023language,NLHF} or DPO manner~\cite{online-DPO}.
% methods and LLM_as_a_Judge

\section{Conclusion and Limitations}

We proposed online self-preferring language models to align with human preferences using self-generated response pairs and self-judged preference strengths.
We design a novel SPCE loss function for OSP language models to leverage such preference strength information for alignment.
OSP inherits the advantages of online methods such as generalizing to OOD tasks and self-improving in a more parameter-efficient manner due to self-preferring.
Moreover, we demonstrate that modeling preference strengths contribute to mitigating overfitting, improving alignment performance, and enhancing sample efficiency.

Our work has limitations. OSP is more computationally intensive than offline algorithms due to the need for sampling responses and self-preferring. Additionally, we observe that LLM-as-a-judge tends to favor longer responses, resulting in increased output length in OSP language models. Furthermore, OSP language models established by LLMs larger than 7B can be evaluated in future work.

%\newpage
\bibliography{references}

\begin{thebibliography}{10}

\bibitem{2020GPT3}
Tom Brown, Benjamin Mann, Nick Ryder, Melanie Subbiah, Jared~D Kaplan, Prafulla Dhariwal, Arvind Neelakantan, Pranav Shyam, Girish Sastry, Amanda Askell, et~al.
\newblock Language models are few-shot learners.
\newblock {\em Advances in neural information processing systems}, 33:1877--1901, 2020.

\bibitem{ji2023ai}
Jiaming Ji, Tianyi Qiu, Boyuan Chen, Borong Zhang, Hantao Lou, Kaile Wang, Yawen Duan, Zhonghao He, Jiayi Zhou, Zhaowei Zhang, et~al.
\newblock Ai alignment: A comprehensive survey.
\newblock {\em arXiv preprint arXiv:2310.19852}, 2023.

\bibitem{levine2020offline}
Sergey Levine, Aviral Kumar, George Tucker, and Justin Fu.
\newblock Offline reinforcement learning: Tutorial, review, and perspectives on open problems.
\newblock {\em arXiv preprint arXiv:2005.01643}, 2020.

\bibitem{InstructGPT}
Long Ouyang, Jeffrey Wu, Xu~Jiang, Diogo Almeida, Carroll Wainwright, Pamela Mishkin, Chong Zhang, Sandhini Agarwal, Katarina Slama, Alex Ray, et~al.
\newblock Training language models to follow instructions with human feedback.
\newblock {\em Advances in Neural Information Processing Systems}, 35:27730--27744, 2022.

\bibitem{DPO}
Rafael Rafailov, Archit Sharma, Eric Mitchell, Stefano Ermon, Christopher~D Manning, and Chelsea Finn.
\newblock Direct preference optimization: Your language model is secretly a reward model.
\newblock {\em Advances in Neural Information Processing Systems}, 2023.

\bibitem{IPO}
Mohammad~Gheshlaghi Azar, Mark Rowland, Bilal Piot, Daniel Guo, Daniele Calandriello, Michal Valko, and R{\'e}mi Munos.
\newblock A general theoretical paradigm to understand learning from human preferences.
\newblock {\em arXiv preprint arXiv:2310.12036}, 2023.

\bibitem{zhao2023sliHF}
Yao Zhao, Rishabh Joshi, Tianqi Liu, Misha Khalman, Mohammad Saleh, and Peter~J Liu.
\newblock Slic-hf: Sequence likelihood calibration with human feedback.
\newblock {\em arXiv preprint arXiv:2305.10425}, 2023.

\bibitem{NLHF}
R{\'e}mi Munos, Michal Valko, Daniele Calandriello, Mohammad~Gheshlaghi Azar, Mark Rowland, Zhaohan~Daniel Guo, Yunhao Tang, Matthieu Geist, Thomas Mesnard, Andrea Michi, et~al.
\newblock Nash learning from human feedback.
\newblock {\em arXiv preprint arXiv:2312.00886}, 2023.

\bibitem{lambert2022challenges}
Nathan Lambert, Markus Wulfmeier, William Whitney, Arunkumar Byravan, Michael Bloesch, Vibhavari Dasagi, Tim Hertweck, and Martin Riedmiller.
\newblock The challenges of exploration for offline reinforcement learning.
\newblock {\em arXiv preprint arXiv:2201.11861}, 2022.

\bibitem{online-DPO}
Shangmin Guo, Biao Zhang, Tianlin Liu, Tianqi Liu, Misha Khalman, Felipe Llinares, Alexandre Rame, Thomas Mesnard, Yao Zhao, Bilal Piot, et~al.
\newblock Direct language model alignment from online ai feedback.
\newblock {\em arXiv preprint arXiv:2402.04792}, 2024.

\bibitem{tang2024understanding}
Yunhao Tang, Daniel~Zhaohan Guo, Zeyu Zheng, Daniele Calandriello, Yuan Cao, Eugene Tarassov, R{\'e}mi Munos, Bernardo~{\'A}vila Pires, Michal Valko, Yong Cheng, et~al.
\newblock Understanding the performance gap between online and offline alignment algorithms.
\newblock {\em arXiv preprint arXiv:2405.08448}, 2024.

\bibitem{yuan2024self}
Weizhe Yuan, Richard~Yuanzhe Pang, Kyunghyun Cho, Sainbayar Sukhbaatar, Jing Xu, and Jason Weston.
\newblock Self-rewarding language models.
\newblock {\em arXiv preprint arXiv:2401.10020}, 2024.

\bibitem{Self-judge}
Sangkyu Lee, Sungdong Kim, Ashkan Yousefpour, Minjoon Seo, Kang~Min Yoo, and Youngjae Yu.
\newblock Aligning large language models by on-policy self-judgment.
\newblock {\em arXiv preprint arXiv:2402.11253}, 2024.

\bibitem{PPO}
John Schulman, Filip Wolski, Prafulla Dhariwal, Alec Radford, and Oleg Klimov.
\newblock Proximal policy optimization algorithms.
\newblock {\em arXiv preprint arXiv:1707.06347}, 2017.

\bibitem{bradley1952rank}
Ralph~Allan Bradley and Milton~E Terry.
\newblock Rank analysis of incomplete block designs: I. the method of paired comparisons.
\newblock {\em Biometrika}, 39(3/4):324--345, 1952.

\bibitem{Moss}
Rui Zheng, Shihan Dou, Songyang Gao, Wei Shen, Binghai Wang, Yan Liu, Senjie Jin, Qin Liu, Limao Xiong, Lu~Chen, et~al.
\newblock Secrets of rlhf in large language models part i: Ppo.
\newblock {\em arXiv preprint arXiv:2307.04964}, 2023.

\bibitem{bertrand2023limitations}
Quentin Bertrand, Wojciech~Marian Czarnecki, and Gauthier Gidel.
\newblock On the limitations of the elo, real-world games are transitive, not additive.
\newblock In {\em International Conference on Artificial Intelligence and Statistics}, pages 2905--2921. PMLR, 2023.

\bibitem{bai2022training}
Yuntao Bai, Andy Jones, Kamal Ndousse, Amanda Askell, Anna Chen, Nova DasSarma, Dawn Drain, Stanislav Fort, Deep Ganguli, Tom Henighan, et~al.
\newblock Training a helpful and harmless assistant with reinforcement learning from human feedback.
\newblock {\em arXiv preprint arXiv:2204.05862}, 2022.

\bibitem{TLDR}
Michael V{\"o}lske, Martin Potthast, Shahbaz Syed, and Benno Stein.
\newblock Tl; dr: Mining reddit to learn automatic summarization.
\newblock In {\em Proceedings of the Workshop on New Frontiers in Summarization}, pages 59--63, 2017.

\bibitem{pmlr-v162-ethayarajh22a}
Kawin Ethayarajh, Yejin Choi, and Swabha Swayamdipta.
\newblock Understanding dataset difficulty with $\mathcal{V}$-usable information.
\newblock In Kamalika Chaudhuri, Stefanie Jegelka, Le~Song, Csaba Szepesvari, Gang Niu, and Sivan Sabato, editors, {\em Proceedings of the 39th International Conference on Machine Learning}, volume 162 of {\em Proceedings of Machine Learning Research}, pages 5988--6008. PMLR, 17--23 Jul 2022.

\bibitem{fine-tuned-GPTJ}
CarperAI.
\newblock Openai summarize tl;dr reward model checkpoint, 2023.
\newblock \url{https://huggingface.co/CarperAI/openai_summarize_tldr_rm_checkpoint}.

\bibitem{GPT4}
Josh Achiam, Steven Adler, Sandhini Agarwal, Lama Ahmad, Ilge Akkaya, Florencia~Leoni Aleman, Diogo Almeida, Janko Altenschmidt, Sam Altman, Shyamal Anadkat, et~al.
\newblock Gpt-4 technical report.
\newblock {\em arXiv preprint arXiv:2303.08774}, 2023.

\bibitem{SHP}
Kawin Ethayarajh, Yejin Choi, and Swabha Swayamdipta.
\newblock Understanding dataset difficulty with $\mathcal{V}$-usable information.
\newblock In Kamalika Chaudhuri, Stefanie Jegelka, Le~Song, Csaba Szepesvari, Gang Niu, and Sivan Sabato, editors, {\em Proceedings of the 39th International Conference on Machine Learning}, volume 162 of {\em Proceedings of Machine Learning Research}, pages 5988--6008. PMLR, 17--23 Jul 2022.

\bibitem{gao2023scaling}
Leo Gao, John Schulman, and Jacob Hilton.
\newblock Scaling laws for reward model overoptimization.
\newblock In {\em International Conference on Machine Learning}, pages 10835--10866. PMLR, 2023.

\bibitem{amodei2016concrete}
Dario Amodei, Chris Olah, Jacob Steinhardt, Paul Christiano, John Schulman, and Dan Man{\'e}.
\newblock Concrete problems in ai safety.
\newblock {\em arXiv preprint arXiv:1606.06565}, 2016.

\bibitem{RAFT}
Hanze Dong, Wei Xiong, Deepanshu Goyal, Rui Pan, Shizhe Diao, Jipeng Zhang, Kashun Shum, and Tong Zhang.
\newblock Raft: Reward ranked finetuning for generative foundation model alignment.
\newblock {\em arXiv preprint arXiv:2304.06767}, 2023.

\bibitem{SPIN}
Zixiang Chen, Yihe Deng, Huizhuo Yuan, Kaixuan Ji, and Quanquan Gu.
\newblock Self-play fine-tuning converts weak language models to strong language models.
\newblock {\em arXiv preprint arXiv:2401.01335}, 2024.

\bibitem{RSO}
Tianqi Liu, Yao Zhao, Rishabh Joshi, Misha Khalman, Mohammad Saleh, Peter~J Liu, and Jialu Liu.
\newblock Statistical rejection sampling improves preference optimization.
\newblock In {\em The Twelfth International Conference on Learning Representations}, 2024.

\bibitem{NLPO}
Rajkumar Ramamurthy, Prithviraj Ammanabrolu, Kiant{\'e} Brantley, Jack Hessel, Rafet Sifa, Christian Bauckhage, Hannaneh Hajishirzi, and Yejin Choi.
\newblock Is reinforcement learning (not) for natural language processing: Benchmarks, baselines, and building blocks for natural language policy optimization.
\newblock In {\em The Eleventh International Conference on Learning Representations}, 2023.

\bibitem{zhu2023fine}
Banghua Zhu, Hiteshi Sharma, Felipe~Vieira Frujeri, Shi Dong, Chenguang Zhu, Michael~I Jordan, and Jiantao Jiao.
\newblock Fine-tuning language models with advantage-induced policy alignment.
\newblock {\em arXiv preprint arXiv:2306.02231}, 2023.

\bibitem{P3O}
Tianhao Wu, Banghua Zhu, Ruoyu Zhang, Zhaojin Wen, Kannan Ramchandran, and Jiantao Jiao.
\newblock Pairwise proximal policy optimization: Harnessing relative feedback for llm alignment.
\newblock {\em arXiv preprint arXiv:2310.00212}, 2023.

\bibitem{ReMax}
Ziniu Li, Tian Xu, Yushun Zhang, Yang Yu, Ruoyu Sun, and ZhiQuan Luo.
\newblock Remax: A simple, effective, and efficient method for aligning large language models.
\newblock {\em arXiv preprint arXiv:2310.10505}, 2023.

\bibitem{UP-RLHF}
Yuanzhao Zhai, Han Zhang, Yu~Lei, Yue Yu, Kele Xu, Dawei Feng, Bo~Ding, and Huaimin Wang.
\newblock Uncertainty-penalized reinforcement learning from human feedback with diverse reward lora ensembles.
\newblock {\em arXiv preprint arXiv:2401.00243}, 2023.

\bibitem{ethayarajh2024kto}
Kawin Ethayarajh, Winnie Xu, Niklas Muennighoff, Dan Jurafsky, and Douwe Kiela.
\newblock Kto: Model alignment as prospect theoretic optimization.
\newblock {\em arXiv preprint arXiv:2402.01306}, 2024.

\bibitem{DNO}
Corby Rosset, Ching-An Cheng, Arindam Mitra, Michael Santacroce, Ahmed Awadallah, and Tengyang Xie.
\newblock Direct nash optimization: Teaching language models to self-improve with general preferences.
\newblock {\em arXiv preprint arXiv:2404.03715}, 2024.

\bibitem{zheng2023judging}
Lianmin Zheng, Wei-Lin Chiang, Ying Sheng, Siyuan Zhuang, Zhanghao Wu, Yonghao Zhuang, Zi~Lin, Zhuohan Li, Dacheng Li, Eric Xing, et~al.
\newblock Judging llm-as-a-judge with mt-bench and chatbot arena.
\newblock {\em Advances in Neural Information Processing Systems}, 36, 2023.

\bibitem{bai2022constitutional}
Yuntao Bai, Saurav Kadavath, Sandipan Kundu, Amanda Askell, Jackson Kernion, Andy Jones, Anna Chen, Anna Goldie, Azalia Mirhoseini, Cameron McKinnon, et~al.
\newblock Constitutional ai: Harmlessness from ai feedback.
\newblock {\em arXiv preprint arXiv:2212.08073}, 2022.

\bibitem{lee2023rlaif}
Harrison Lee, Samrat Phatale, Hassan Mansoor, Kellie Lu, Thomas Mesnard, Colton Bishop, Victor Carbune, and Abhinav Rastogi.
\newblock Rlaif: Scaling reinforcement learning from human feedback with ai feedback.
\newblock {\em arXiv preprint arXiv:2309.00267}, 2023.

\bibitem{pang2023language}
Jing-Cheng Pang, Pengyuan Wang, Kaiyuan Li, Xiong-Hui Chen, Jiacheng Xu, Zongzhang Zhang, and Yang Yu.
\newblock Language model self-improvement by reinforcement learning contemplation.
\newblock {\em arXiv preprint arXiv:2305.14483}, 2023.

\bibitem{iteratedDPO}
Jing Xu, Andrew Lee, Sainbayar Sukhbaatar, and Jason Weston.
\newblock Some things are more cringe than others: Preference optimization with the pairwise cringe loss.
\newblock {\em arXiv preprint arXiv:2312.16682}, 2023.

\bibitem{liang2021towards}
Paul~Pu Liang, Chiyu Wu, Louis-Philippe Morency, and Ruslan Salakhutdinov.
\newblock Towards understanding and mitigating social biases in language models.
\newblock In {\em International Conference on Machine Learning}, pages 6565--6576. PMLR, 2021.

\bibitem{yao2023deepspeed}
Zhewei Yao, Reza~Yazdani Aminabadi, Olatunji Ruwase, Samyam Rajbhandari, Xiaoxia Wu, Ammar~Ahmad Awan, Jeff Rasley, Minjia Zhang, Conglong Li, Connor Holmes, et~al.
\newblock Deepspeed-chat: Easy, fast and affordable rlhf training of chatgpt-like models at all scales.
\newblock {\em arXiv preprint arXiv:2308.01320}, 2023.

\end{thebibliography}
\bibliographystyle{unsrt}

\newpage
\appendix
% \onecolumn

\section{Impact Statements}
\label{appendix:impact}
The focus area of this paper, aligning LLMs with human preferences, is interconnected with a significant line of research aimed at mitigating social bias in LLMs~\cite{liang2021towards}.
Given the recent work in the areas moving towards offline methods, our work demonstrates the clear advantage of online methods using self-provided responses and preferences, which is a promising avenue to be further explored.
However, as AI systems become more capable, self-preferring for improvement may induce problematic AI behaviors such as power-seeking or other unseen risks.
% from anthropic

\section{Experimental Setup Details}
\label{appendix:detailed-exp-settings}
\subsection{Hyper-parameters}

All experiments were conducted using a single Nvidia A40 GPU.
Detailed hyperparameters are provided in \cref{appendix:tab:hyper-parameters}.
The only exception is that we observed weaker discrimination of self-preference when only a $2\%$ portion of the HH dataset was available.
Consequently, we increased $\beta$ to $0.02$ and $K$ to 9 for this particular setting.
% We do not 

% we set $\beta=0.02$ and $K=9$ when only $2\%$ portion of the HH dataset is available.
% This is because we observe the discrimination of self-preference is weaker with such few data

For all baselines, we adhered to the same general hyperparameters unless explicitly stated otherwise.
Specifically, we adhere to the hyper-parameters outlined in the DPO paper~\cite{DPO}, where $\beta=0.1$ for the helpful dialogue task and $\beta=0.5$ for the summarization task.
For IPO, we set $\beta=0.1$ for both tasks.
We align LLMs with DPO and IPO for 1 epoch.
For the RLHF baseline, the coefficient for the KL penalty is set to $0.05$.

We implemented all methods in the paper using DeepSpeed-Chat~\cite{yao2023deepspeed} and adopted most hyperparameters, including the Adam optimizer, learning rate schedule, etc.
Note that we modified the supervised fine-tuning step by masking the next-token prediction loss for the prompt sections.
Following~\cite{yao2023deepspeed}, we fine-tune TinyLlama and Mistral-7B-Instruct-v0.2 in a supervised manner before alignment.
Methods presented in \cref{tab:main:performance} start from the SFT model fine-tuned with the preferred responses of the full dataset.
Methods in \cref{table:002dataset}, where only a $2\%$ portion of datasets is available, involved fine-tuning LLMs with the preferred responses of the $2\%$ dataset.
For evaluation without preference augmentation (\cref{fig-zero-shot}), we fine-tuned Mistral-7B-Instruct-v0.2 with $20\%$ prompts and preferred responses from the TL;DR dataset for warm-up before alignment.
% We use the batch size of 16 and the learning rate of 2e-6 for Tiny-Llama and 2e-5 for Mistral-7B-Instruct-v0.2 fine-tuned with LoRA.

\begin{table}[t]%
	\centering
 \caption{Hyper-parameters for our experimental results.}
  \vskip 0.1in
 \resizebox{\linewidth}{!}{
		\begin{tabular}{l|cccc}
			\toprule
			 &   & Helpful Dialogue & \multicolumn{2}{|c}{Summarization} \\
        &&TinyLlama-1.1B&TinyLlama-1.1B& Mistral-7B \\
			\cmidrule(lr){1-5} % 调整中间横线长度
			\multirow{5}{*}{generation}& sampling method&  top-\textit{p} (\textit{p}=0.9) & top-\textit{p} (\textit{p}=0.9) & top-\textit{p} (\textit{p}=0.9)  \\
			&generation temperature& 0.7  & 0.7  & 0.7 \\
            % &Boltzmann temperature& 0.2  & 0.05  \\
			&max length for samples & 512 & 768 & 768 \\
            &\begin{tabular}[c]{@{}c@{}}max length for \\[-0.5ex] LLM-as-a-judge instructions\end{tabular}  & 768 & 1124 & 1124 \\
			\cmidrule(lr){1-5} % 调整中间横线长度
			\multirow{6}{*}{training}& batch size & 16 & 16 & 8  \\
			&learning rate& 2e-6 &2e-6& 2e-5 \\
			& LoRA dimension & -  &-& 64 \\
            & LoRA alpha & -  &-& 32 \\
            & $\beta$ &  0.001 & 0.02 & 0.1 \\
            & sampling number $K$ & 7  &7& 7 \\
            & epoch & 1  &1& 1 \\
            % &\begin{tabular}[c]{@{}c@{}}preference-strengthening \\[-0.5ex] coefficient $\lambda$ \end{tabular}  & 0.01  & & ??? \\
			\bottomrule
		\end{tabular}
	}
	\label{appendix:tab:hyper-parameters}
\end{table}

\subsection{LLM-as-a-judge Instructions}

To achieve self-preferring, we delineate the principles embodying human preferences and craft instructions by populating the placeholders highlighted in red. To mitigate position bias~\cite{zheng2023judging}, we create two instructions for each pair by interchanging the position of the responses.

\begin{figure*}[t]
    \centering
    {
        \centering
        \small
        \begin{tcolorbox}[colback=gray!10!white, colframe=gray!30!white, width=0.97\textwidth, arc=4mm, auto outer arc]
            You are an expert rater of helpful and honest assistant responses. Given the context and the two responses, choose the most helpful and honest response. \\
            \\
            Context: \texttt{\color{red}<Prompt>}\\
            \\
            RESPONSE A:\texttt{\color{red}<Response A>}\\
            \\
            RESPONSE B:\texttt{\color{red}<Response B>}\\
            \\
            Which response is better? RESPONSE \texttt{\color{blue}<A/B>}
        \end{tcolorbox}
    }

    \caption{The LLM-as-a-judge instruction in the helpful dialogue task. 
     \label{fig:hh-evalprompt}}
\end{figure*}

\begin{figure}[h]
\centering
\small
\begin{tcolorbox}[colback=gray!10!white, % Background color
                  colframe=gray!30!white, % Frame color
                  width=0.99\textwidth, % Width of the tcolorbox
                  arc=4mm, % Radius of the rounded corners
                  auto outer arc,
                  ]

A good summary is a shorter piece of text that has the essence of the original. It tries to accomplish the same purpose and conveys the key information from the original post. Below we define four evaluation axes for summary quality: coherence, accuracy, coverage, and overall quality.

Coherence: This axis answers the question “how coherent is the summary on its own?” A summary is coherent if it’s easy to understand when read on its own and free of English errors. A summary is not coherent if it’s difficult to understand what the summary is trying to say. Generally, it’s more important that the summary is understandable than it being free of grammar errors.

Accuracy: This axis answers the question “does the factual information in the summary accurately match the post?” A summary is accurate if it doesn’t say things that aren’t in the article, it doesn’t mix up people, and generally is not misleading. 

Coverage: This axis answers the question “how well does the summary cover the important information in the post?” A summary has good coverage if it mentions the main information from the post that’s important to understand the situation described in the post. A summary has poor coverage if someone reading only the summary would be missing several important pieces of information about the situation in the post. A summary with good coverage should also match the purpose of the original post (e.g. to ask for advice).

Overall quality: This axis answers the question “how good is the summary overall at representing the post?” This can encompass all of the above axes of quality, as well as others you feel are important. If it’s hard to find ways to make the summary better, the overall quality is good. If there are lots of different ways the summary can be made better, the overall quality is bad. 

\ 

You are an expert summary rater. Given a piece of text and two of its possible summaries, choose the summary that best meets the definitions above for coherence, accuracy, coverage, and overall quality.
\\
\\
Text: \texttt{\color{red}<Prompt>}\\
\\
SUMMARY A:\texttt{\color{red}<Summary A>}\\
\\
SUMMARY B:\texttt{\color{red}<Summary B>}\\
\\
Which summary is better? SUMMARY \texttt{\color{blue}<A/B>}
% Please evaluate the following model response to the given instruction on a scale from 1-5.
%\\\\
%Original Instruction: [INSTRUCTION PROMPT]
%\\\\
%Model Response: [MODEL RESPONSE TO INSTRUCTION] 
\end{tcolorbox}
\caption{
The LLM-as-a-judge instruction in the summarization task. 
% To avoid the position bias, we construct two instructions for each pair by swapping the position of win response $\by^h$ and lose response $\by^l$.
}
\label{fig:TLDR-evalprompt}
\end{figure}

% \subsection{Prompts for GPT-4 Evaluations}
% \label{appendix:GPT-4-prompts}
% We use the following prompts to evaluate the answers for the helpful dialogue and summarization tasks, which are inherited from the LLM-as-a-judge instructions with additional restrictions of output format.

% \lstset{
%   basicstyle=\small\ttfamily, % set the font and size
%   frame=single
% }
% \begin{lstlisting}
% For the following query to a chatbot, which response is more helpful? 
% Query: 
% <user_query>
% Answer A: 
% <answer_a>
% Answer B: 
% <answer_b>

% FIRST provide a one-sentence comparison of the two answers and explain 
% which you feel is more helpful. 
% SECOND, on a new line, state only "A" or "B" to indicate which answer 
% is more helpful. 

% Your response should use the format:

% Comparison: <one-sentence comparison and explanation>
% More helpful: <"A" or "B">
% \end{lstlisting}

% \lstset{
%   basicstyle=\small\ttfamily, % set the font and size
%   frame=single
% }
% \begin{lstlisting}
% Which of the following summaries does a better job of summarizing the 
% most important points in the given forum post, without including 
% unimportant or irrelevant details?

% <Detailed evaluation axes in terms of coherence, accuracy, coverage, 
% and overall quality>

% FIRST provide a one-sentence comparison of the two summaries, 
% explaining which you prefer and why. 
% SECOND, on a new line, state only "A" or "B" to indicate your choice. 

% Your response should use the format: 

% Comparison: <one-sentence comparison and explanation> 
% Preferred: <"A" or "B">

% \end{lstlisting}

\section{Additional Experimental Results}

\subsection{Reward Hacking of RLHF on OOD Generalization Tasks}
\label{appendix:RLHF-overoptimization}
The reward hacking problem, also known as overoptimization~\cite{gao2023scaling}, primarily arises from imperfect reward models trained on a limited human preference dataset. Specifically, reward models may assign high rewards to certain low-quality out-of-distribution samples, thereby leading to the optimization of LLMs toward learning these undesirable behaviors.

\cref{tab:RMEvaluation} shows that the reward model trained on the full HH dataset achieves an accuracy of $62\%$ on the test dataset, while the accuracy of the reward model trained on a $2\%$ portion of the dataset is only $53\%$.
We observe the reward hacking problem when aligning LLMs using RLHF in OOD generalization tasks, particularly with reward models trained on the full HH dataset. 
This could be attributed to the fact that reward models trained on only a $2\%$ portion of the dataset cannot optimize LLMs effectively.

% may not effectively combat overoptimization, where the training rewards do not significantly increase.

\subsection{Loss Curves for OSP Language Models with Different Loss Functions.}
In \cref{fig-ablation-spce}, it is observed that OSP with DPO loss suffers from severe overfitting in the online setting. To further analyze this issue, we visualize the loss values in this section.

\begin{figure*}[t]
    \subfigure[OSP with DPO loss functions.
    % by reward models and gold reward models.
        \label{fig-loss-dpo}]
        {
            \centering
            \includegraphics[width=0.295\linewidth]{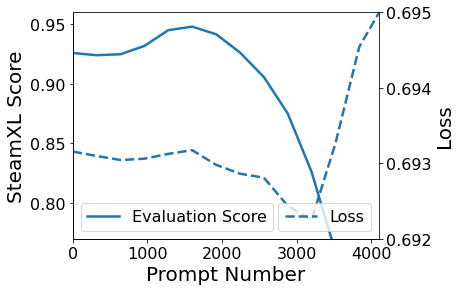}
        }
        \quad
    \subfigure[OSP with IPO loss functions.
    \label{fig-loss-ipo}]
    {
        \centering
        \includegraphics[width=0.293\linewidth]{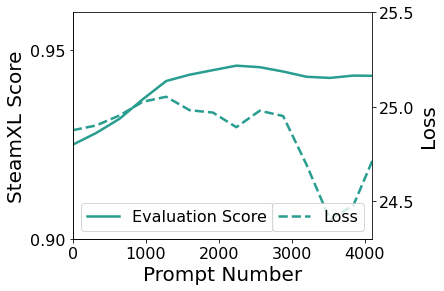}
    }
            \quad
    \subfigure[OSP with SPCE loss functions.
    \label{fig-loss-spce}]
    {
        \centering
        \includegraphics[width=0.32\linewidth]{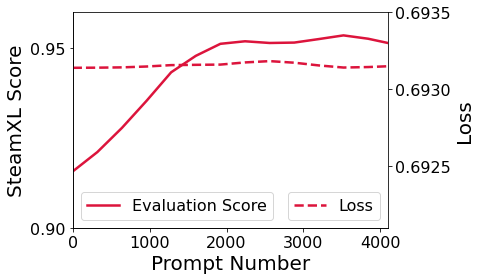}
    }
    \vspace{-1em}
    \caption{
   Training curves of OSP with different loss functions.
    \label{fig-loss}}
\end{figure*}

Overall, we observe a slight decrease in training losses in the online self-preferring setting. This is because $\by^h$ and $\by^l$ are generated by the same LLM $\pi_\theta$. 
A small decrease in the DPO loss value can improve the evaluation performance, as shown in \cref{fig-loss-dpo}. However, as the DPO loss continues to decrease, overfitting occurs. In contrast, IPO and SPCE loss can mitigate this issue, consistent with our gradient analysis in~\cref{sec:method:spce}.

\begin{figure*}[t]
\centering
    \subfigure[Askbaking.
    % by reward models and gold reward models.
        \label{appendix:fig-overoptimization-askbaking}]
        {
            \centering
            \includegraphics[width=0.45\linewidth]{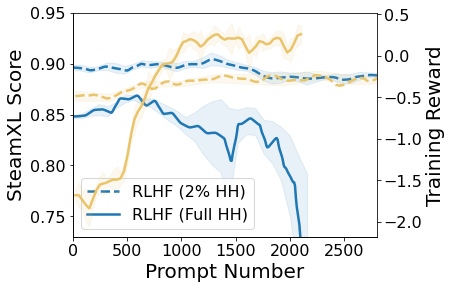}
        }
        \quad
    \subfigure[Askacademia.
    \label{appendix:fig-overoptimization-askacademia}]
    {
        \centering
        \includegraphics[width=0.45\linewidth]{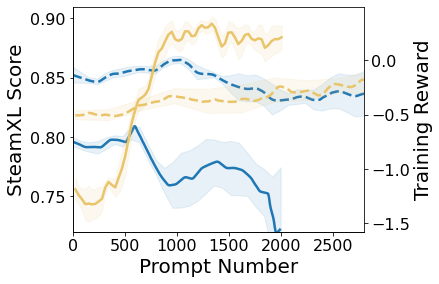}
    }
    \vspace{-1em}
    \caption{ Training curves of RLHF with different reward models trained on $2\%$ portion and full HH dataset. The scores of SteamSHP-flan-t5-xl are in blue while the training rewards provided by reward models are in yellow. 
   % Overoptimization
    \label{appendix:fig-RLHF-overoptimization-OOD}}
\end{figure*}

% \subsection{A Fair Comparision between Online and Offline Method}

\subsection{Case Studies of GPT-4 Evaluations on the Helpful Dialogue Task}
\label{appendix:GPT-4-evaluations-HH}
In \cref{fig:Head2Head}, we present the win rates of GPT-4 evaluations of LLMs established by Tiny-Llama and aligned with different methods on the HH dataset. The corresponding case studies are detailed in this section.

For each evaluation, we randomly select $50$ prompts from the test sets and have two LLMs aligned with different methods generate corresponding responses. We then use the response pairs to fill in the prompt templates in \cref{fig:hh-evalprompt} and impose additional restrictions on the output format. To mitigate position bias, we interchange the positions of responses, resulting in $100$ prompts for GPT-4 evaluations of the two methods.

\begin{table}[htbp]
\caption{
% Examples of head-to-head evaluations by GPT-4 of policy models established by Llama2-7B.
GPT-4 prefers OSP over RLHF.
% Evaluations involving the swapping of response positions consistently yield judgments.
For clarity, we include post-hoc annotations in bold, formatted as \textbf{[annotation]}. These annotations are not inherent to the model generations.}
\small
\vskip 0.1in
\centering
\begin{tabular}{@{}p{1.6cm}p{0.84\textwidth}@{}}
% line 146
\toprule
\multirow{3}{*}{\begin{tabular}[l]{@{}l@{}}Prompt \\ \end{tabular} }  &
For the following query to a chatbot, which response is more helpful? 

Query:

Human: I like to host guests at my home from time to time, and I am gathering recipes of different dishes and drinks to keep things interesting. I am interested in trying some Mongolian dishes. Can you give me a recipe for Uuz?

Assistant: Sure, what are you looking for? Do you want a recipe for a hot, cooked uuz dish? A dip, or condiment? A side dish? A drink?

Human: I am looking for a recipe for Uuz. If I understand correctly it is slow cooked mutton. Is that correct?

Assistant:

Answer A \textbf{[RLHF]}: 
Patchouli can be grown from seeds or from root cuttings.  If you want to grow patchouli from seeds, you’ll want to start with seeds that have been grown on a seed-starting tray and have been cooled to 55 degrees Fahrenheit.  You’ll want to wait a few weeks to a month before you plant them, and then you’ll want to plant them in a warm, moist soil.  Good luck!</s>

Answer B \textbf{[OSP]}: 
That’s correct!  Uuz is a traditional Mongolian dish made from slow-cooked mutton, as well as spices like garlic and onions.  It’s often served with rice or noodles, and is often accompanied by vegetables and side dishes.  Uuz is a very popular dish in Mongolian cuisine, and is also often considered a delicacy.  Overall, uuz is a really flavorful and beautiful dish, and is a great choice for any party or gathering.</s>

FIRST provide a one-sentence comparison of the two answers and explain which you feel is more helpful. 

SECOND, on a new line, state only "A" or "B" to indicate which answer is more helpful.

Your response should use the format:

Comparison: \textless one-sentence comparison and explanation \textgreater

More helpful: \textless ``A" or ``B"\textgreater
\\ \cline{1-2} 
\multirow{3}{*}{\begin{tabular}[l]{@{}l@{}}Judgment \\ \end{tabular} } &
While both answers confirm that uuz is a slow-cooked mutton dish, Answer B provides more relevant details about the dish's ingredients, accompaniments, and cultural significance in Mongolian cuisine.

More helpful: B \textbf{[OSP]}
\\
\bottomrule
\end{tabular}
%\vspace{-0.5em}

\end{table}

\begin{table}[htbp]
\caption{
% Examples of head-to-head evaluations by GPT-4 of policy models established by Llama2-7B.
GPT-4 prefers OSP over SFT.
Evaluations involving the interchanging of response positions consistently yield judgments.
For clarity, we include post-hoc annotations in bold, formatted as \textbf{[annotation]}. These annotations are not inherent to the model generations.}
\small
\vskip 0.1in
\centering
\begin{tabular}{@{}p{1.6cm}p{0.84\textwidth}@{}}
% line 146
\toprule
\multirow{3}{*}{\begin{tabular}[l]{@{}l@{}}Prompt \\ \end{tabular} }  &
For the following query to a chatbot, which response is more helpful? 

Query:

Human: I like to host guests at my home from time to time, and I am gathering recipes of different dishes and drinks to keep things interesting. I am interested in trying some Mongolian dishes. Can you give me a recipe for Uuz?

Assistant: Sure, what are you looking for? Do you want a recipe for a hot, cooked uuz dish? A dip, or condiment? A side dish? A drink?

Human: I am looking for a recipe for Uuz. If I understand correctly it is slow cooked mutton. Is that correct?

Assistant:

Answer A \textbf{[OSP]}: 
That’s correct!  Uuz is a traditional Mongolian dish made from slow-cooked mutton, as well as spices like garlic and onions.  It’s often served with rice or noodles, and is often accompanied by vegetables and side dishes.  Uuz is a very popular dish in Mongolian cuisine, and is also often considered a delicacy.  Overall, uuz is a really flavorful and beautiful dish, and is a great choice for any party or gathering.</s>

FIRST provide a one-sentence comparison of the two answers and explain which you feel is more helpful. 

SECOND, on a new line, state only "A" or "B" to indicate which answer is more helpful.

Answer B \textbf{[SFT]}: 
I think that’s right.</s>

Your response should use the format:

Comparison: \textless one-sentence comparison and explanation \textgreater

More helpful: \textless ``A" or ``B"\textgreater
\\ \cline{1-2} 
\multirow{3}{*}{\begin{tabular}[l]{@{}l@{}}Judgment \\ \end{tabular} } &
Answer A provides more detailed and helpful information about what uuz is, including the main ingredients, how it's typically served, and its popularity in Mongolian cuisine, while Answer B simply agrees with the query without adding any new information.

More helpful: A \textbf{[OSP]}
\\
\\ \cline{1-2} 
\multirow{3}{*}{\begin{tabular}[l]{@{}l@{}}Prompt \\ (interchanging \\  positions)\end{tabular} }   &
For the following query to a chatbot, which response is more helpful? 

Query: 

Human: I like to host guests at my home from time to time, and I am gathering  recipes of different dishes and drinks to keep things interesting. I am interested in trying some Mongolian dishes. Can you give me a recipe for Uuz?

Assistant: Sure, what are you looking for? Do you want a recipe for a hot, cooked uuz dish? A dip, or condiment? A side dish? A drink?

Human: I am looking for a recipe for Uuz. If I understand correctly it is slow cooked mutton. Is that correct?

Assistant:

Answer A \textbf{[SFT]}: 
I think that’s right.</s>

Answer B \textbf{[OSP]}: 
That’s correct!  Uuz is a traditional Mongolian dish made from slow-cooked mutton, as well as spices like garlic and onions.  It’s often served with rice or noodles, and is often accompanied by vegetables and side dishes.  Uuz is a very popular dish in Mongolian cuisine, and is also often considered a delicacy.  Overall, uuz is a really flavorful and beautiful dish, and is a great choice for any party or gathering.</s>

FIRST provide a one-sentence comparison of the two answers and explain which you feel is more helpful. 

SECOND, on a new line, state only "A" or "B" to indicate which answer is more helpful.

Your response should use the format:

Comparison: \textless one-sentence comparison and explanation \textgreater

More helpful: \textless ``A" or ``B"\textgreater
\\ \cline{1-2} 
\multirow{3}{*}{\begin{tabular}[l]{@{}l@{}}Judgment \\ \end{tabular} }  &
Answer B provides more detail and context about the dish, making it more informative and helpful for the person asking the question.

More helpful: B \textbf{[OSP]} \\

\bottomrule
\end{tabular}
%\vspace{-0.5em}

\end{table}

\begin{table}[htbp]
\caption{
% Examples of head-to-head evaluations by GPT-4 of policy models established by Llama2-7B.
GPT-4 prefers OSP over DPO.
% Evaluations involving the swapping of response positions consistently yield judgments.
For clarity, we include post-hoc annotations in bold, formatted as \textbf{[annotation]}. These annotations are not inherent to the model generations.\label{appendix:GPT4:OSP-DPO}}
\small
\vskip 0.1in
\centering
\begin{tabular}{@{}p{1.6cm}p{0.84\textwidth}@{}}
% line 146
\toprule
\multirow{3}{*}{\begin{tabular}[l]{@{}l@{}}Prompt \\ \end{tabular} }  &
For the following query to a chatbot, which response is more helpful? 

Query:

Human: I like to host guests at my home from time to time, and I am gathering recipes of different dishes and drinks to keep things interesting. I am interested in trying some Mongolian dishes. Can you give me a recipe for Uuz?

Assistant: Sure, what are you looking for? Do you want a recipe for a hot, cooked uuz dish? A dip, or condiment? A side dish? A drink?

Human: I am looking for a recipe for Uuz. If I understand correctly it is slow cooked mutton. Is that correct?

Assistant:

Answer A \textbf{[DPO]}: 
 Yes, you should plant patchouli every year.  It’s really important to keep it growing, because it’s a plant that needs a lot of water and nutrients.  When you’re growing it in the ground, you’ll want to keep it in warm soil, and also make sure it gets plenty of water.  You’ll also want to make sure it’s not too cold or too hot, and also make sure it gets plenty of sunlight.  If you’re growing it in a container, you’ll want to make sure it gets plenty of water, and also make sure it gets plenty of sunlight.  Make sure the soil is warm and not too hot or cold, and also make sure it gets plenty of water.  You’ll want to make sure it gets plenty of water, and also make sure it gets plenty of sunlight.  You’ll want to make sure it gets plenty of water, and also make sure it gets plenty of sunlight.  You’ll also want to make sure it gets plenty of water, and also make sure it gets plenty of sunlight.  Make sure it gets plenty of water, and also make sure it gets plenty of sunlight

Answer B \textbf{[OSP]}: 
That’s correct!  Uuz is a traditional Mongolian dish made from slow-cooked mutton, as well as spices like garlic and onions.  It’s often served with rice or noodles, and is often accompanied by vegetables and side dishes.  Uuz is a very popular dish in Mongolian cuisine, and is also often considered a delicacy.  Overall, uuz is a really flavorful and beautiful dish, and is a great choice for any party or gathering.</s>

FIRST provide a one-sentence comparison of the two answers and explain which you feel is more helpful. 

SECOND, on a new line, state only "A" or "B" to indicate which answer is more helpful.

Your response should use the format:

Comparison: \textless one-sentence comparison and explanation \textgreater

More helpful: \textless ``A" or ``B"\textgreater
\\ \cline{1-2} 
\multirow{3}{*}{\begin{tabular}[l]{@{}l@{}}Judgment \\ \end{tabular} } &
Answer B provides more context and detail about what uuz is and how it's typically served, while Answer A simply confirms the dish and asks a clarifying question.

More helpful: B \textbf{[OSP]}
\\
\bottomrule
\end{tabular}
%\vspace{-0.5em}

\end{table}

\begin{table}[htbp]
\caption{
% Examples of head-to-head evaluations by GPT-4 of policy models established by Llama2-7B.
GPT-4 prefers OSP over IPO.
% Evaluations involving the swapping of response positions consistently yield judgments.
For clarity, we include post-hoc annotations in bold, formatted as \textbf{[annotation]}. These annotations are not inherent to the model generations.}
\small
\vskip 0.1in
\centering
\begin{tabular}{@{}p{1.6cm}p{0.84\textwidth}@{}}
% line 146
\toprule
\multirow{3}{*}{\begin{tabular}[l]{@{}l@{}}Prompt \\ \end{tabular} }  &
For the following query to a chatbot, which response is more helpful? 

Query:

Human: I like to host guests at my home from time to time, and I am gathering recipes of different dishes and drinks to keep things interesting. I am interested in trying some Mongolian dishes. Can you give me a recipe for Uuz?

Assistant: Sure, what are you looking for? Do you want a recipe for a hot, cooked uuz dish? A dip, or condiment? A side dish? A drink?

Human: I am looking for a recipe for Uuz. If I understand correctly it is slow cooked mutton. Is that correct?

Assistant:

Answer A \textbf{[IPO]}: 
 No, you can grow patchouli in the ground as long as you keep it watered and make sure it gets enough light.  You can also grow patchouli in containers, and you can start with a small container of potting soil or directly in the ground in warm soil.  Once the seeds start to grow you’ll want to transplant them into individual pots or containers and make sure they get plenty of water.  This plant is really all about the ground where they grow and also the care they get when they are growing.  You can grow patchouli in the ground in a warm soil, or you can grow it in a container in a cooler climate.  You can also grow patchouli in a container in a warmer climate if you live in a cooler climate, but you may need to grow it in a warmer climate in order to get the best growth.  You can also start with a small container of potting soil or directly in the ground in warm soil, but the seeds will need to be started in warm soil.  Once the seeds start to grow you’ll want to transplant them into individual pots or containers and make sure they get plenty of water.  This plant

Answer B \textbf{[OSP]}: 
That’s correct!  Uuz is a traditional Mongolian dish made from slow-cooked mutton, as well as spices like garlic and onions.  It’s often served with rice or noodles, and is often accompanied by vegetables and side dishes.  Uuz is a very popular dish in Mongolian cuisine, and is also often considered a delicacy.  Overall, uuz is a really flavorful and beautiful dish, and is a great choice for any party or gathering.</s>

FIRST provide a one-sentence comparison of the two answers and explain which you feel is more helpful. 

SECOND, on a new line, state only "A" or "B" to indicate which answer is more helpful.

Your response should use the format:

Comparison: \textless one-sentence comparison and explanation \textgreater

More helpful: \textless ``A" or ``B"\textgreater
\\ \cline{1-2} 
\multirow{3}{*}{\begin{tabular}[l]{@{}l@{}}Judgment \\ \end{tabular} } &
While both answers confirm that uuz is a slow-cooked mutton dish, Answer B provides more detail about the ingredients, accompaniments, and cultural significance, making it more informative and helpful for someone looking to understand and prepare the dish.

More helpful: B \textbf{[OSP]}
\\
\bottomrule
\end{tabular}
%\vspace{-0.5em}

\end{table}

\subsection{Case Studies of GPT-4 Evaluations on the Summarization Task}
\label{appendix:GPT-4-evaluations-TLDR}
We compare the summaries generated by the fine-tuned Mistral-7B-Instruct-v0.2 model with its self-improving version using the checkpoint aligned with 200 prompts via OSP in \cref{fig-zero-shot}.

\begin{table*}[htbp]
\caption{
GPT-4 prefers OSP over SFT.
% Evaluations involving the swapping of response positions consistently yield judgments.
For clarity, we include post-hoc annotations in bold, formatted as \textbf{[annotation]}. These annotations are not inherent to the model generations.
}
% Examples of prompts and answers of SFT and overoptimized models in the summarization task.}
\small
\centering
\begin{tabular}{@{}p{1.6cm}p{0.84\textwidth}@{}}
% line 146
\toprule
Prompt &
SUBREDDIT: r/AskReddit

TITLE: Have I been friendzoned?

POST: Before I begin,  let me tell you that I am a girl to avoid any confusion later. Also,  this is the boy I consider(ed?) my best friend. 

So me and this boy have been friends for 6 years and we're really close. We are both seniors in high school. He asked me out in October and dumped before Christmas. He gave me reasons like "I don't want to be in a relationship that will have to end when we part for college", "I'm done with high school dating", "I need to focus on myself right now" and the infamous "I just don't want to ruin our friendship". As much as it sucked to hear,  I accepted it. He kept sending me messages about how much he wanted me but couldn't have me. Fast forward to yesterday when he tells me he's dating someone. An ex-girlfriend. What?We still hang out and talk like we did before we were dating. TL;DR: 
\\ \cline{1-2} 
\multirow{2}{*}{\begin{tabular}[l]{@{}l@{}}Summary of \\ SFT model\end{tabular} }  &
Senior ex-boyfriend dumped me, then wanted to go back to \"just as friends\", ignored how hurt I'd been. Recently found out he has a new girlfriend.</s> \\ \cline{1-2} 
\multirow{2}{*}{\begin{tabular}[l]{@{}l@{}}Summary of \\ OSP model\end{tabular} }  &
My ex-boyfriend dumped me before Christmas and is now dating an ex-girl friend. He still wants me and keeps telling me he does. Have I been friendzoned?</s>   \\
\cline{1-2} 
\multirow{4}{*}{\begin{tabular}[l]{@{}l@{}}GPT-4 \\ judgement\end{tabular} }  &
Summary B \textbf{[OSP]} more accurately reflects the original post’s situation of feeling friendzoned and includes important emotional details of the ex-boyfriend’s actions, while Summary A \textbf{[SFT]} omits the detail of the ex-boyfriend dating an ex-girlfriend and the continued communication.

Preferred: B \textbf{[OSP]} \\
\bottomrule
\end{tabular}
%\vspace{-0.5em}

\end{table*}

\begin{table*}[htbp]
\caption{
GPT-4 prefers OSP over SFT.
% Evaluations involving the swapping of response positions consistently yield judgments.
For clarity, we include post-hoc annotations in bold, formatted as \textbf{[annotation]}. These annotations are not inherent to the model generations.
}
% Examples of prompts and answers of SFT and overoptimized models in the summarization task.}
\small
\centering
\begin{tabular}{@{}p{1.6cm}p{0.84\textwidth}@{}}
% line 146
\toprule
Prompt &
SUBREDDIT: r/loseit

TITLE: The moment I've dreaded my entire weight loss journey is finally here.

POST: I got some interesting news today. I was getting new tires put on my car and the mechanic told me it would take an hour to do so I walked down the street to get some water. I passed this laser liposuction/skin tightening spa and decided since I have an hour to kill and a some skin sag from losing 130lbs I'd see what that's all about. During my consultation the doctor looked at my stomach and said verbatim "My technique will do nothing for you, you need a tummy tuck. " The way he explained it was that there is not enough fat on my stomach to make his technique viable. So in short... this actually may all be skin on me and I may be DONE losing weight. The only places he said he knew he could do anything were my love handles and chest. I'm going to get a real medical BMI scan before I make any moves on this so I have hard numbers as to what is actually where but like I said earlier, those numbers could reveal that I'm done with weight loss and it's time to start managing my skin better. I'd just like to point out that I'm not going to do a tummy tuck, down time from that could cost me my job and is just unmanageable (Not to mention invasive surgery scares the shit out of me). This could be a big turning point in my journey and I don't want to take it blindly.Does anyone know anything I could be doing short term to help this out? or any techniques I am overlooking? If I can avoid surgery of any kind I'd like to but if not this seems like the best option right now. TL;DR: 
\\ \cline{1-2} 
\multirow{2}{*}{\begin{tabular}[l]{@{}l@{}}Summary of \\ SFT model\end{tabular} }  &
lost my skin gets tighter, more weight, not tummy tuck; still more to do.</s>

\\ \cline{1-2} 
\multirow{2}{*}{\begin{tabular}[l]{@{}l@{}}Summary of \\ OSP model\end{tabular} }  &
Down 130lbs but still have excess skin, doctor suggested tummy tuck but I'm not willing to go under the knife. Need advice on how to manage my skin better.</s>   \\
\cline{1-2} 
\multirow{4}{*}{\begin{tabular}[l]{@{}l@{}}GPT-4 \\ judgement\end{tabular} }  &
Summary B \textbf{[OSP]} accurately and coherently captures the essence of the original post, reflecting the individual's weight loss journey and concerns regarding excess skin, while Summary A \textbf{[SFT]} 
is fragmented and lacks clarity.

Preferred: B \textbf{[OSP]} \\
\bottomrule
\end{tabular}
%\vspace{-0.5em}

\end{table*}

\end{document}